\documentclass[10pt,twocolumn,letterpaper]{article}

\usepackage{wacv}              % To produce the CAMERA-READY version
\usepackage{graphicx}
\usepackage{amsmath}
\usepackage{amssymb}
\usepackage{booktabs}
\usepackage[accsupp]{axessibility}

\usepackage[pagebackref,breaklinks,colorlinks]{hyperref}

\usepackage[capitalize]{cleveref}
\crefname{section}{Sec.}{Secs.}
\Crefname{section}{Section}{Sections}
\Crefname{table}{Table}{Tables}
\crefname{table}{Tab.}{Tabs.}

\usepackage{eccvabbrv}

\usepackage{graphicx}
\usepackage{booktabs}

\usepackage{amsmath}
\usepackage{amssymb}
\usepackage{booktabs}

\usepackage{times}
\usepackage{epsfig}
\usepackage{enumitem}
\usepackage{diagbox}
\usepackage{threeparttable}
\usepackage[table]{xcolor}
\usepackage{multirow}
\usepackage{xcolor}
\usepackage{colortbl}
\newcommand{\tildea}[1]{\overset{\sim}{#1}}

\def\wacvPaperID{863} % *** Enter the WACV Paper ID here
\def\confName{WACV}
\def\confYear{2025}

\begin{document}


% \title{Classifying the State of Anything \\  Zero-Shot Object-agnostic State Classification}
% \title{Anything State Classification \\  Zero-Shot Object-agnostic State}
% \title{State of Anything\\  Zero-Shot Object-agnostic State Classification}
\title{Recognizing Unseen States of Unknown Objects \\ by  Leveraging Knowledge Graphs} %:Leveraging Knowledge Graphs for Zero-Shot Object-agnostic State Classification}

% \author {
%     %not Authors
%     Filippos Gouidis\textsuperscript{\rm 1,\rm 2,\rm 3},
%     Konstantinos Papoutsakis\textsuperscript{\rm 2}
%     Theodore Patkos\textsuperscript{\rm 3}, \\
%     Antonis Argyros\textsuperscript{\rm 2,\rm  3} and
%     Dimitris Plexousakis\textsuperscript{\rm 2,\rm  3}
% }
% \affiliations {
%     % Affiliations
%   \textsuperscript{\rm 1}Computer Science Department, University of Crete, Heraklion, Greece
  
%     \textsuperscript{\rm 2}Department of Management, Science and Technology, Hellenic Mediterranean University, Agios Nikolaos, Greece \\
%     \textsuperscript{\rm 3}Institute of Computer Science, Foundation for Research and Technology Hellas, Heraklion, Greece \\
%          gouidis@\{ics.forth.gr, csd.uoc.gr, hmu.gr\}, kpapoutsakis@hmu.gr, \{papanton, patkos, argyros, dp\}@ics.forth.gr
% }

\author{Filippos Gouidis\textsuperscript{\rm 1,\rm  2}
% Foundation for Research and Technology Hellas,\\
% Heraklion, Greece \\
\and
Konstantinos Papoutsakis\textsuperscript{\rm 3}
\and
Theodore Patkos\textsuperscript{\rm 1}
\and
Antonis Argyros\textsuperscript{\rm 1,\rm  2}
\and
Dimitris Plexousakis\textsuperscript{\rm 1,\rm  2}\\
   \textsuperscript{\rm 1} Foundation for Research and Technology-Hellas,  Greece \\
\textsuperscript{\rm 2} University of Crete, Greece 
\textsuperscript{\rm 3} Hellenic Mediterranean University, Greece\\
{\tt\small  \{gouidis,patkos,argyros,dp\}@ics.forth.gr, kpapoutsakis@hmu.gr}
}

% \author{Filippos Gouidis\\
% Institute of Computer Science, Foundation for Research and Technology Hellas, Heraklion, Greece 

% {\tt\small gouidis@ics.forth.grmember  student}
% % \and
% %     Konstantinos Papoutsakis

% % \and
% %       Theodore Patkos

% % \and
% %     Konstantinos Papoutsakis

% % \and
% %     Dimitris Plexousakis
% }
\maketitle

%%%%%%%%% ABSTRACT
\begin{abstract}
We investigate the problem of Object State Classification (OSC) in the context of zero-shot learning. Specifically, we propose the first method for Zero-shot Object-agnostic State Classification (OaSC)  that, given an image, infers the state of a single object without relying on the knowledge or the estimation of the object class. In that direction, we capitalize on Knowledge Graphs (KGs) for structuring and organizing external knowledge, which, in combination with visual information, enable effective inference of the states of objects that have not been encountered in the training set. 
%This work presents  the first zero-shot method for object-agnostic OSC. 
Having this unique property, a significant strength of our method is that it can handle an Open Set of object classes. 
We investigate the performance of OaSC in various datasets and settings, against several hypotheses and in comparison with state-of-the-art approaches for object attribute classification. OaSC outperforms these methods significantly across all benchmarks.\footnote{Code and models are publicly available at {\url{https://github.com/philipposg/OaSC.git}.}} %OaSC outperforms these methods in all datasets and benchmarks by an important margin. %Moreover, the experimental results demonstrate that the leveraging of KG knowledge of an object class is not decisive for the prediction of its state. 
% Our code is publicly available\footnote{\url{https://anonymous.4open.science/r/Exps_Code__863_WACV2025-ECEC/}.}.
\end{abstract}

\vspace{-0.5cm}

\section{Introduction}

In our daily lives, we interact with objects regularly for various purposes and in various contexts, \textcolor{black}{
often bringing changes in object states. The object state change can be seen as the effect of the transformation induced by the interaction~\cite{wang2016actions}. %%%XXX: This change can be either temporary (i.e. a bottle of water is opened) or permanent/irreversible (i.e. an apple was chopped into slices) and may affect the appearance, shape, or functionality of the object.
}
% In our daily lives, we interact with objects regularly for various purposes and in various contexts.
%to use them in various goal-oriented contexts in workplaces, domestic, and other environments. 
% Objects may be in different states.
% that can be defined as a subset of perceptible attributes~\cite{Isola2015} that are closely related to potential changes in the appearance, shape, and functionality of objects as a precondition for or an effect of an action (e.g., unfolded, closed, full etc.)~\cite{liu2017jointly}. 
The recognition of object states and state changes is crucial for determining an object's condition and the interaction that was performed or a future one the object could afford~\cite{jamone2016affordances}. These cues highlight the significance of the \textcolor{black}{Object State Classification (OSC) task in computer vision that can leverage the functionality and performance of AI systems in tasks such as learning object affordances~\cite{chuang2018learning}, recognizing interactions~\cite{Wang2016b,Isola2015,liu2017jointly,Mancini2022}, reasoning to achieve an object state change~\cite{Farhadi2009}, recognizing the completion or failure of goals,  recovery from possible mistakes~\cite{schoonbeek2024industreal}, etc.}

%%%XXX during procedural activities

% In Computer Vision, object state classification (OSC) is an important task 
%for inferring an object's functionality 
% and is closely related to action recognition~\cite{Wang2016b}, object detection and classification~\cite{Farhadi2009} and affordance learning~\cite{chuang2018learning}. 

Despite the importance of OSC, the amount of research on this problem is notably limited, particularly when compared with the research on the related area of object classification.
% Given the importance of OSC, the research on state classification is disproportionately low, especially in comparison to the enormous effort that has been invested on the related field of object classification. 
However, this seems to have changed during the last few years as the number of works dedicated to this problem keeps growing~\cite{Isola2015,gouidis2022,Soucek2022,10376814}. \textcolor{black}{Large-scale video datasets~\cite{Grauman_2022_CVPR,10376814} of human-object interactions now offer rich annotation data related to object state changes and define new problems and establish benchmarks and challenges related to object state detection and classification~\cite{Grauman_2022_CVPR}.}

\begin{figure}[t]
\centering
\includegraphics[width=\columnwidth]{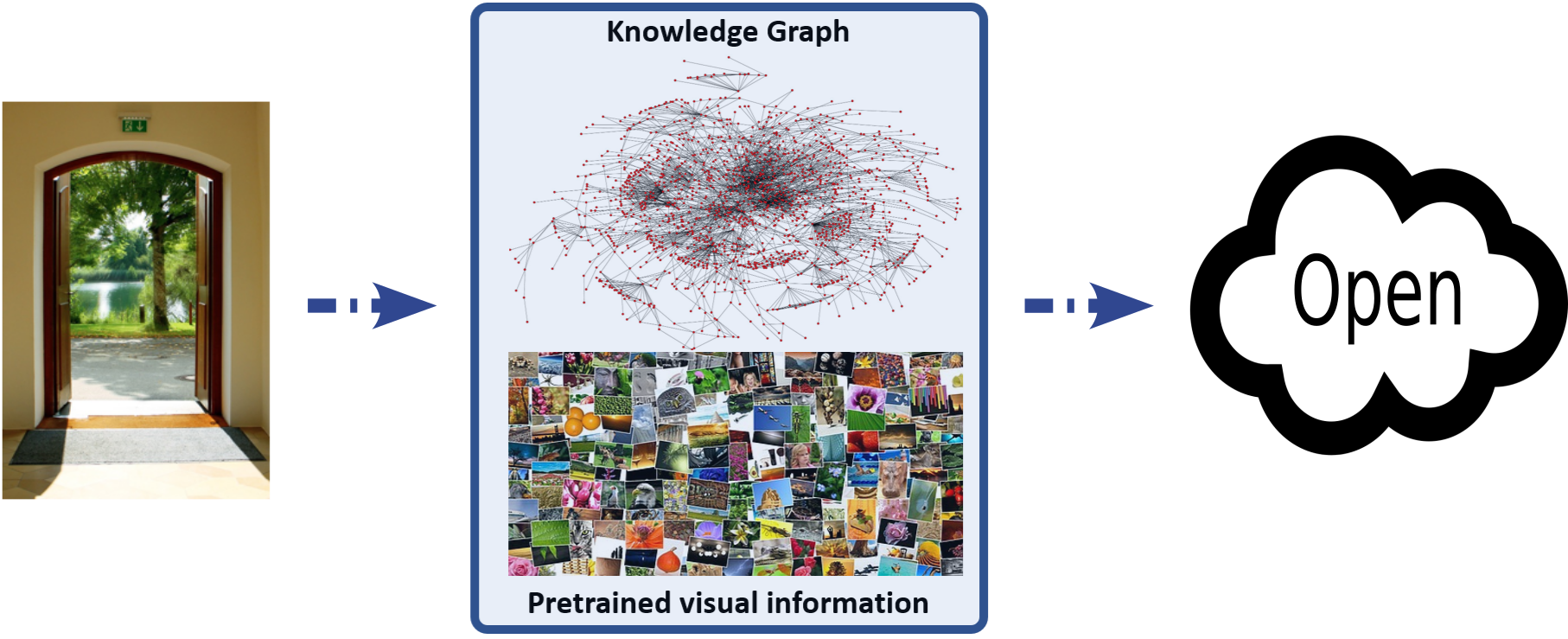}\\
%\caption{The proposed method for Object-agnostic State Classification (OaSC) combines (a)~structured knowledge on object states stemming from common-sense knowledge repositories with (b)~visual information that relates the appearance of certain objects to their states. Some of the  objects states classes might have training samples, but, in general, the method can cope with state classes without training samples. Regarding the classes of the objects our method does not depend on any training samples and  enabling inference of the state of objects that have never been presented to the training set. For example, a door can be inferred as open, even if no doors are present in the training set.}

%%%AAA below the version I propose. Please, check!!!
\caption{The proposed method for Object-agnostic State Classification (OaSC) combines (a)~structured knowledge on object states stemming from common-sense knowledge repositories with (b)~visual information related to seen object state classes. % (known at training time). 
By leveraging these information sources, OaSC can classify the state for any object, regardless of its class, i.e. object-agnostic classification, and can also infer new state classes that are not seen in the training set.
%visual information that is object-agnostic and corresponds to state classes.
%that relates the appearance of certain objects to their states. 
%The proposed method can estimate object states which may have never been seen at training time. Moreover, since it does not rely on information about object classes it can also estimate the states of novel objects. 
For example, a door can be inferred as open, even if the training set contains no doors and no other open objects.}

%This combination enables inference of the state of objects that have never been presented to the training set.}
\label{imag:ex1}
%\vspace{-0.65cm} 
\end{figure}

In the context of visual object recognition, %states can also be viewed as a distinct subset of perceptible object attributes. 
states represent a unique subset of perceptible object attributes.
Attributes typically refer to static visual or other types of properties of objects, such as color, shape, or texture. In contrast, states are defined based on changes in appearance or context, which are more subtle and can be influenced by various factors. Moreover, states provide cues on the dynamic aspects and transformation of an object's physical and/or functional properties as a result of actions. 
% This information is also crucial towards accurate action recognition in images or videos~\cite{Isola2015,liu2017jointly,Mancini2022}. 
% Recognizing states in images is generally more challenging compared to attributes due to the complexity involved in representing visual information. 
% Attributes are typically defined based on visual properties that remain relatively stable across different contexts and appearances. In contrast, states are defined based on changes in appearance or context, which are more subtle and can be influenced by various factors.
Therefore, accurately recognizing states poses challenges such as capturing and modeling the dynamic nature of visual information, identifying subtle changes in appearance, and accounting for contextual variations across all possible objects that can be seen in each specific state.% (see \autoref{fig:fig2}). 
\textcolor{black}{To tackle these challenges, we seek inspiration from the notion and techniques of compositional learning and zero-shot classification~\cite{Saini_2022_CVPR} to attempt disentanglement of objects and the
states classes in images. In essence, we focus on learning prototypical representations of state classes regardless of the object classes to capture state-specific features of, e.g. anything open, closed, plugged, etc, in an open-world setting. 
% with improved generalization and cross-domain applicability against large image sets containing both known and new, previously unknown object categories
} %TODO: SHOULD WE MENTION OPEN-WORLD BUZZWORD HERE OR NOT?

Towards this end, we investigate a zero-shot variant for the OSC problem (see \autoref{imag:ex1}) by focusing on images containing household objects. Specifically, we developed and extensively evaluated a novel zero-shot object-agnostic State Classification method (OaSC) that does not rely on object class-related information. 
% (see~\autoref{imag:ex1}). 
Our approach explores the potential benefits of Knowledge Graphs (KGs) as a well-established, powerful tool for structuring and organizing external knowledge that can be applied to various fields, including zero-shot learning. We argue that KGs can enhance the accuracy and robustness of models for the OSC task as they provide structured representations of the relationships among different entities and concepts, enabling the inference of relationships among unseen and seen/known categories.  
The proposed method is the first zero-shot approach that focuses on this problem enabling the recognition of states of previously unseen object classes. Despite its potential practical merits, such a feature is currently not supported by zero-shot attribute classification methods. 
% This variation of OSC may provide important practical merits; i.e., enables a robotic helper to recognize the states of unknown objects. 

%The main advantage of the zero-shot object-aware methods is that the correct classification of an object class can help the recognition of its state. However, in cases where the object class is misclassified the accurate state classification becomes much more challenging if not impossible. The state classification for any object-aware method can either take place after the object class recognition or in parallel. The first approach is not computationally feasible, since it entails the utilization of different state classifiers for each object class. The second approach is less computationally expensive and is followed by most of the object-aware methods currently. A major drawback in this case is that in real-world scenarios when the valid object-state combinations are known beforehand the search space for the classifiers corresponds to the Cartesian product of the objects and the states classes which can be very large. Equally important, these methods require samples of every object and state class during the training, since they are zero-shot w.r.t.  object-state pairs and not w.r.t. states or the object classes per se. Therefore, in contrast to our method, these methods are not tailored to address state classification in an open-world setting for the object classes. 

Zero-shot object-aware methods excel in classifying object classes to facilitate state recognition, yet struggle when an object class is misidentified, making state classification difficult. These methods operate in two stages (classifying the object first and then its state) or in one stage (doing both simultaneously). 
%, which is computationally intensive, or by doing both simultaneously, which is more common and less demanding. 
In both cases, a major limitation is the expansive search space for classifiers in real scenarios, driven by a large set of combinations of object and state classes. Furthermore, such methods require training samples for all object and state classes, making them unsuitable for state classification that is open w.r.t. object classes, unlike object-agnostic approaches like our method.
Consider, for example, a scenario where $500$ different object classes can be situated in $20$ different states. If a two-stage object-aware method is used, 500 different state classifiers should be trained, whereas, in the case of a one-stage object-aware method, the classifier has to consider the $10,000$ labels of all the object/state pairs. In contrast, by following our approach, we employ a single classifier that considers the space of $20$ state labels.
%Besides, it is important to note that since the number of state classes is much more limited than that of object classes it is much easier to know them in advance.
%household objects' states and does not rely on object classification. This aspect of our approach enables the recognition of states in classes of objects that are not known beforehand, a property that current zero-shot attribute classification does not support. Our contributions can be summarized as follows:

Overall, our contributions can be summarized as follows:

\begin{itemize}[noitemsep,topsep=0pt]
%\begin{itemize}
\itemsep0em 
\item We introduce the problem of object-agnostic zero-shot state classification and we propose OaSC, a new method for solving it. In contrast to object-informed zero-shot methods, OaSC does not rely on prior accurate object classification, exhibiting thus greater robustness and applicability.
% We propose a novel zero-shot state classification method that outperforms existing state-of-the-art methods. 
% Notably, our approach is object-agnostic, i.e., its performance does not rely on prior accurate object classification, resulting in greater robustness than other competing methods. To the best of our knowledge, our method is the first of its kind. 
% Notably, our approach is object-agnostic, i.e., its performance does not rely on prior accurate object classification, resulting in greater robustness than other competing methods. To the best of our knowledge, our method is the first of its kind. 
\item 
An extensive experimental evaluation is conducted across 4 datasets and 11 state-of-the-art compositional zero-shot learning methods. Our method achieves a performance that is superior by a great margin.
\item  The ablation study reported explores the strengths and weaknesses of our proposed method in various settings. This analysis provides valuable insights related to the new problem and method.
%into the new OaSC problem towards developing further improvements.

\end{itemize}

\section{Related Work}

\noindent\textbf{State/Attribute Classification}:
The most generally accepted definition of ``visual attributes" refers to visual concepts that are detectable by machines and can be comprehended by humans~\cite{duan2012discovering}.
The current approach for learning attributes in images is similar to that of object classes, where a convolutional neural network is trained with discriminative classifiers using annotated image datasets~\cite{singh2016end}. However, labeled attribute image datasets often lack the data scale found in object datasets, contain a limited number of generic attributes, or cover only a few specific categories~\cite{lampert2009learning, Isola2015,patterson2016coco, yu2017semantic, Mancini2022}. %Most of these works are based on the same assumptions that are used for the task of attribute classification, with 
%the number of works focusing exclusively on state classification being limited~\cite{gouidis2022, DBLP:conf/visigrapp/GouidisPPAP24}. 
% \vspace*{0.2cm}\noindent\textbf{Open-vocabulary Attribute Detection} 
% \vspace*{0.2cm}\noindent\textbf{Zero-shot Learning} 
% Zero-Shot Learning (ZSL) is a task that involves recognizing new classes that were not seen during training, by leveraging side information such as attributes\cite{Lampert2014}, text descriptions\cite{reed2016learning} or word embeddings \cite{socher2013zero}. Previous approaches have attempted to learn a compatibility function between image and class embeddings\cite{akata2013label} or generate image features for novel classes \cite{xian2018feature}.
Few studies  address explicitly state classification~\cite{gouidis2022,DBLP:conf/visigrapp/GouidisPPAP24}, with most adopting assumptions from attribute classification. Zero-shot learning has emerged as a prominent approach, leveraging semantic embeddings for object representation~\cite{Wang2018b}, and recent works integrate Knowledge Graphs (KGs) or combine KGs with Large Language Models (LLMs)~\cite{gouidis2024llm,DBLP:conf/aaaiss/GouidisPPPAP24}. Other methods focus on compositional image generation~\cite{saini2023chop} or conditioned diffusion models for object state transformations~\cite{soucek2024genhowto}. In the context of videos object state changes provide meaningful context for video-based human action recognition (HAR), complementing visual action representations. Methods often detect object states explicitly~\cite{fathi2013modeling, soucek2022look} or indirectly via scene changes~\cite{alayrac2017joint}. Notable works include frameworks for discovering object states and manipulation actions~\cite{alayrac2017joint}, modeling object fluents in egocentric videos~\cite{Liu2017}, and analyzing multi-object interactions~\cite{Ma2018}. Recent methods leverage self-supervised learning for temporal localization~\cite{soucek2022look}, open-world object part segmentation~\cite{xue2024learning},  disentangling embeddings for object-state recognition~\cite{saini2022disentangling} and anticipation of object states changes~\cite{manousaki2024anticipating}.%, with models like InternVideo~\cite{chen2022ego4d} adapted for object state change and action anticipation.

\noindent\textbf{Zero-shot Object Classification}:
Zero-shot object classification has gained increasing attention due to its practical importance in real-world applications, where it is often difficult to obtain training data for all possible object classes~\cite{xian2018zero}. Several approaches were proposed to address this problem, including semantic embedding-based methods~\cite{Wang2018b, xian2018feature,fu2015zero}, attribute-based methods~\cite{Lampert2014},  generative models~\cite{xian2018feature,changpinyo2016synthesized} and learning of a compatibility function between image and class embeddings ~\cite{akata2015evaluation}. Semantic embedding-based methods employ compact semantic spaces or attribute sets to bridge seen and unseen object classes. Attribute-based methods leverage a set of attributes that describe object classes and use these attributes to infer the class of an unseen object. Generative models generate samples of unseen object classes by synthesizing images that are similar to images of seen object classes. In addition to these approaches, recent work has explored the use of knowledge graphs~\cite{Kampffmeyer2019,nayak:tmlr22}, which capture semantic relationships between objects and can be used to facilitate zero-shot learning. Prior methods in zero-shot learning utilized predetermined attributes or pretrained embeddings, in contrast to our approach which centers on acquiring class representations directly from the knowledge graph during the task.
In a similar vein, some recent works~\cite{DBLP:conf/aaaiss/GouidisPPPAP24,gouidis2024llm} have explored the role of Large Language Models (LLMs) in the context of zero-shot classification.
% Overall, these approaches have achieved promising results in zero-shot object classification and hold great potential for real-world applications such as text classification, video action recognition, and machine translation. 
% There has been significant research on zero-shot object classification using graph neural networks, with recent works exploring the use of common sense knowledge graphs to generate class representations. However, previous methods relied on predefined attributes or pretrained embeddings, while our approach focuses on explicitly learning class representations from the knowledge graph in the task. Other notable works in zero-shot learning include text classification, video action recognition, and machine translation.  

\noindent\textbf{Compositional Zero-shot Learning}: Compositional Zero-shot Learning (CZSL) aims to generalize to unseen combinations of object and state primitives by learning compositionality from the training set. Approaches are grouped into two types: one models individual classifiers for states and objects or learns hierarchical visual primitives~\cite{misra2017red, nagarajan2018attributes, yang2020learning, karthik2022open}, while the other learns a joint compatibility function between image, state, and object~\cite{purushwalkam2019task,atzmon2020causal}. For instance, \cite{atzmon2020causal} introduced a causal graph ensuring primitive independence, while \cite{Li2020} used a symmetry-based framework inspired by group theory. Graph CNNs were employed by~\cite{Mancini2022} to model dependencies and estimate composition feasibility. More recent works explore disentanglement and external knowledge integration, such as ConceptNet for predicting primitives~\cite{karthik2022open}, generative models for creating novel compositions~\cite{li2022siamese}, and attribute-object invariant domains~\cite{zhang2022learning}. Others focus on learning conditional attribute embeddings~\cite{wang2023learning} or disentangled embeddings via cross-attentions~\cite{hao2023learning}. A key limitation in existing CZSL methods is their reliance on training samples containing attribute-object labels. By contrast, our method models states object-agnostically, enabling generalization to unseen state classes.

\noindent\textbf{Graph Neural Networks}:
Graph Neural Networks (GNNs) have gained popularity due to their ability to learn node embeddings that reflect the structure of the graph \cite{kipf2016semi}. These networks have shown significant improvements in downstream tasks, such as node classification and graph classification \cite{hamilton2017inductive,wu2019simplifying, shang2019end,vashishth2019composition}. In this work, we use the GNN transformers that have recently been used for zero-shot object classification~\cite{nayak:tmlr22}. Prior works have considered transformers as a method to learn meta-paths in heterogeneous graphs rather than as a neighborhood aggregation technique~\cite{yun2022graph,liu2023meta}.  Furthermore, GNNs have been applied to various problems including fine-grained entity typing~\cite{xiong2019imposing}, text classification~\cite{yao2019graph}, reinforcement learning~\cite{adhikari2020learning} and neural machine translation~\cite{bastings2017graph}.

% Graph Convolutional Networks (GCN) \cite{} are a special type of neural networks that leverage the interdependencies of data (nodes) that are defined in a graph. However, current methods \cite{}  have limitations in terms of network depth as over-smoothing at deeper layers of the network can cause all nodes to converge to the same value \cite{} . To address this limitation, several approaches have been proposed, such as dense skip connections \cite{} , randomly dropping edges [\cite{} , and applying a linear combination of neighbor features \cite{} . Recent works in this direction have combined residual connections with identity mapping \cite{} or utilized GCNs for zero-shot learning. 

% For instance, \cite{}  propose to directly predict the classifier weights of novel classes using a GCN applied on an external knowledge graph such as WordNet \cite{} . \cite{}  enhance this approach by introducing a dense graph to learn a shallow GCN as a solution for the Laplacian smoothing problem.

\vspace*{0.0cm}\noindent\textbf{Common Sense Knowledge Graphs}: 
Common sense KGs have been extensively utilized in various tasks including transductive zero-shot text classification \cite{zhang2019integrating} and object classification \cite{Kampffmeyer2019,xian2018zero}. Works such as~\cite{bhagavatula2019abductive} and~\cite{bosselut2019comet}  have explored the application of common sense KGs in diverse settings. The work in~\cite{zhang2019integrating}  used ConceptNet~\cite{speer2017conceptnet} for transductive zero-shot text classification as shallow features for class representation. 
Another work~\cite{zhang2019tgg} also utilized common sense knowledge graphs and GNNs for transductive zero-shot object classification. This approach learns to model seen-unseen relations with a graph neural network and requires knowledge of unseen classes during training, utilizing hand-crafted attributes.   Drawing inspiration from~\cite{nayak:tmlr22} which proposed a novel GNN architecture capable of generating dense vector representations from ConceptNet, we extend this approach in a novel context.

% \vspace*{0.0cm}\noindent\textbf{Large Pre-trained Models}: 
% While Large Language Models (LLMs) have achieved remarkable success in natural language processing tasks, their application to vision tasks has been limited. A recent development concerns the development of LLMs that are adapted to the visual domain. These models, which are typically called Large Pre-trained Models (LPMs), have been trained on massive amounts of text and visual data and exploit contrastive learning in order to understand relationships between these two modalities. This approach enables the utilization of LPMs in downstream tasks such as zero-shot classification, visual question answering and image captioning. Two prominent examples of LPMs are  \cite{radford2021learning} and \cite{jia2021scaling}. 

% We draw inspiration from  where  a novel GNN architecture  is proposed capable of generating dense vector representations from ConceptNet. We extend this approach further and use it a novel context. 

% In contrast, our method does not require explicit knowledge of unseen classes during training, but  learns instead class representations from the common sense knowledge graph. Overall, common sense knowledge graphs have shown great potential in improving the performance of various tasks, and our work extends this by applying them to a novel task.which

\section{Methodology}
\label{sec:method}
\begin{figure*}[t]
    \centering
\noindent\includegraphics[width=0.9\textwidth]{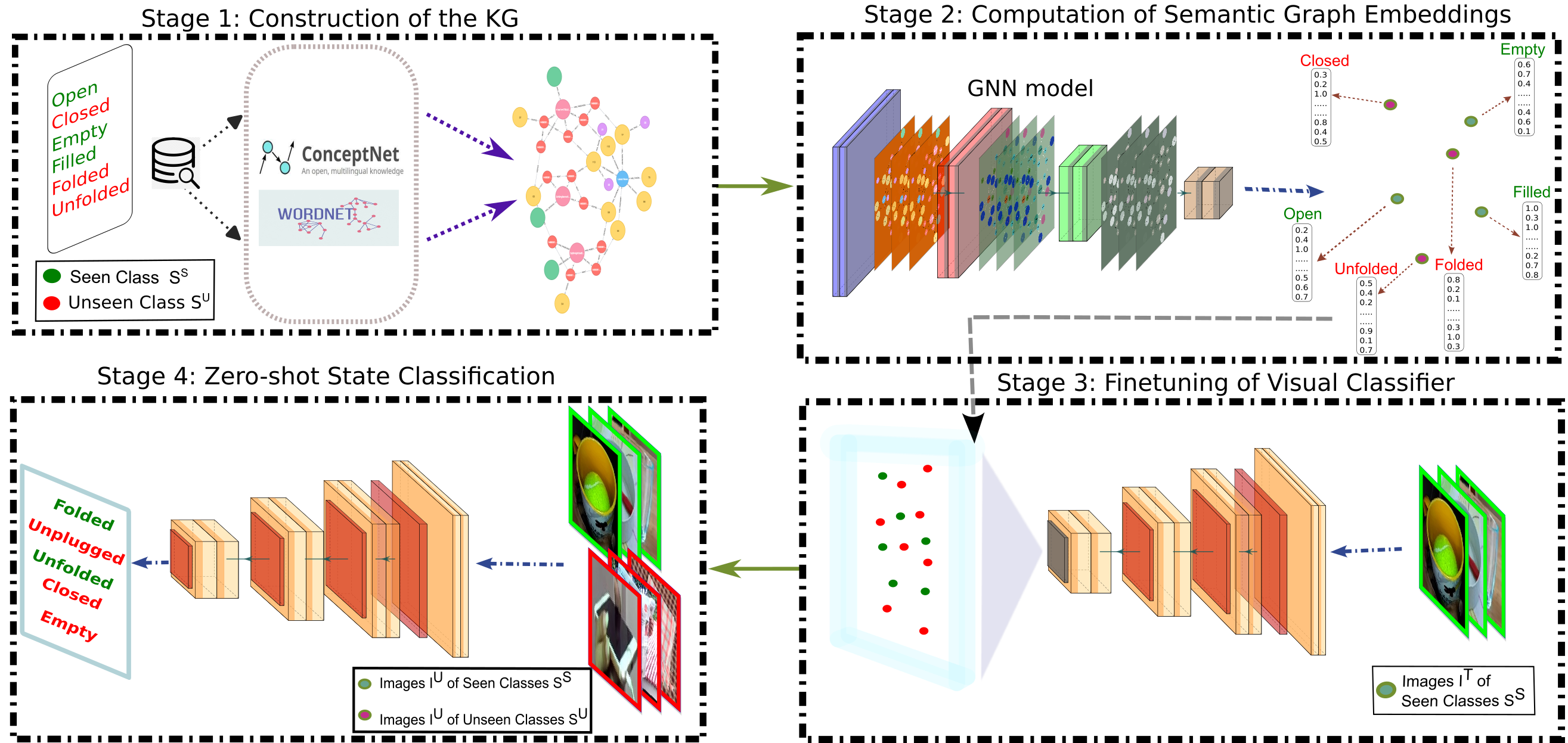}
\caption{The pipeline of OaSC. Our method consists of four stages. In  Stage 1, using as reference points the concepts of seen and unseen state classes (referring to state classes that appear and do not appear in the training set of images, respectively),  a common-sense repository is queried for a KG to be constructed. In Stage 2, the KG is processed by a GNN, which computes embeddings for all state classes (both seen and unseen). These embeddings serve as the final layer of a pre-trained classifier (a CNN model). In Stage 3, the classifier is fine-tuned using images that only contain seen classes, with the last layer of the classifier being fixed. Finally, in Stage 4, the fine-tuned classifier can be utilized for prediction in images including both types of state classes.}

    \label{fig:pipeline}
     % \vspace{-0.5cm} 
    %  \vspace{-0.5cm} 

\end{figure*}

Let $O$ denote a set of objects, $S$ denote the set of states and $I$ denote the set of images, which is partitioned into the training set $I^T$ and the testing set $I^U$. Each image $i \in I$ contains an object $o \in O$ in a state $s \in S$. 
The goal of OSC is to predict the state $s \in S$, given the object $o$ in $i \in I^U$. In the zero-shot variation of OSC, the set of states observed in the test images $S^U$ is not a subset of the set of states observed in the training images $S^S$, i.e., there exists some states in the test image set that do not appear in the training set. Furthermore, the task should be addressed in an object-agnostic manner, i.e. no information concerning the object classes is to be utilized explicitly.  However,  although the set of object classes does not directly affect the task of OaSC, its size is proportional to the complexity of the problem. 
The workflow of the proposed method is shown in \autoref{fig:pipeline}.

% and will be analyzed in the following sections.

% \vspace*{0.2cm}\noindent\textbf{Approach.}

\subsection{Overview}
% Our method is inspired by works that address the problem of zero-shot object classification \cite{}. The main idea behind this line of work is that the necessary information for the classification of the unseen classes can be found in a Knowledge Graph (KG) if processed appropriately by a Graph Neural Network (GNN). Obviously, the most crucial component of this approach lies in the combination of the visual information stemming from the training images and referring to the seen classes with the semantic information stemming from the KG  and referring to the unseen classes.

We are inspired by prior research on zero-shot object classification and leverage the potential of KGs and GNNs to classify previously unseen objects~\cite{Kampffmeyer2019,nayak:tmlr22}. 
The core idea is that semantic information that is stored in the KG can be used by GNNs to learn graph embeddings that can be utilized jointly with visual information extracted from training images. 
This enables the model to generalize to new object classes by leveraging the semantic and contextual information encoded in the graph embeddings of the KG.

% More in detail, the GNN architecture is adopted to the architecture of the Classifier  that is used for the training on seen classes, the GNN last layer has the same size  with the Classifier last layer. This way the GNN can produce semantic embedding features that correspond to all the classes, both seen and unseen, that will be encountered during the inference. These embedding features  replace the last layer of the Classifier. Holding this layer fixed, the body of the Classifier is then fine-tuned with the training images.

GNNs are designed to operate on graph-structured data, such as KGs~\cite{kipf2016semi,Monka2022}. KGs are typically represented as labeled multi-graphs, where nodes correspond to entities, and edges represent entity relationships. GNNs process this graph by iteratively aggregating information from neighboring nodes, using neural network-based operations.

At each iteration, a GNN receives a feature vector for each graph node, which is initially set to the node's embedding vector. Then, the GNN performs a message-passing step that aggregates information from neighboring nodes, based on the edge weights and the features of the nodes. This message-passing operation can be formulated as a neural network layer, which applies a learnable function to the features of the neighboring nodes and returns an aggregated message for each node. After the message-passing step, the GNN updates the node features by applying a learnable transformation that takes into account the original features of the node and the received messages from its neighbors. This updated feature vector is then passed to the next iteration of the message-passing step. The process continues until a fixed number of epochs or convergence.
%%%AAA: Endexetai na mas rethrown gia tis times aytwn twn parametrwn?
% KP edw anaferetai genika mia diadikasia GNN training. Na anaferoume edw times parametrwn h sto 4 - see implementation details ?

The proposed method leverages GNN training using a visual classifier that is trained on seen state classes as supervision. In particular, the last layer of the GNN is designed to have the same size as the last layer of the classifier. This enables the GNN to generate semantic embedding features that correspond to all classes, including both seen and unseen classes that will be encountered during inference. Subsequently, the semantic embedding features replace the last layer of the classifier while this layer is kept fixed. The body of the classifier is then fine-tuned with the training images to optimize the overall model for state recognition.

% \vspace*{0.2cm}\noindent\textbf{GNN Details.} 
Overall, we experimented with four different model architectures and opted for the Transformer Graph Convolutional network (Tr-GCN)~\cite{nayak:tmlr22}. Further details are provided in Section~\ref{sec:abl} and the supplementary material of this work. 
The Tr-GCN mode is capable of combining input sets non-linearly by utilizing multilayer perceptrons and self-attention. Tr-GCN refers to an inductive model that can learn node representations by aggregating local neighborhood features allowing the trained model to make predictions on new graph structures without retraining. 
We leverage the aforementioned property of the Tr-GCN to train a permutation invariant non-linear aggregator that captures the intricate structure of a common sense knowledge graph. 
% , rendering it well-suited for zero-shot learning. 
% It is worth noting that a similar network architecture has been effectively employed for zero-shot object classification~\cite{nayak:tmlr22}.

% A critical aspect of the proposed method involves calibrating the weights of the GNN in a manner that its predictions in the semantic space are useful for the classifier deployed in the visual space. To accomplish this, we adopt an approach based on prior research \cite{Kampffmeyer2019, Wang2018b, nayak:tmlr22} that involves learning the semantic class representations by minimizing the L2 distance between the learned class representations and the weights of a fully connected layer in a ResNet classifier pre-trained on the ILSVRC 2012 dataset \cite{russakovsky2015imagenet}. Once the class representations are learned, we fix them and fine-tune the ResNet backbone using the training images from the dataset.

% \vspace*{0.2cm}\noindent\textbf{Building of the KG.}
% The KG is created by the querying  of a common sense repository. The repositories that we are ConceptNet \cite{} and WordNet\cite{}. The procedure takes place as follows. Initially we create a set of nodes that correspond to the target stace classes. Subsequently, the repository is queried for each of these nodes and its neighbours in the repository of  added to the KG if  certain criteria are met (see ablation section for more details). This procedure is repeated for the newly added nodes and henceforth until a number of hops has been reached.  

\subsection{The proposed OaSC approach}
\label{sec:pipeline}
Overall, the proposed method consists of four stages, as shown in \autoref{fig:pipeline}: (1) construction of the KG, (2) GNN training and learning of semantic graph embeddings, (3) fine-tuning of the visual classifier and (4) deployment of the fine-tuned state classifier.

\vspace*{0.0cm}\noindent\textbf{Construction of the KG (Stage 1)}:
To create the KG, we query a common sense repository to compile a generic solution and to avoid the construction of a task-specific KG, tailored to the entities at hand and their relationships. First, a set of nodes that correspond to the words of the target state classes $S^U$ and $S^S$ is generated. Then, we query the repository for each of these nodes and add their neighbors in the KG, if they meet specific criteria (see also Section~\ref{sec:abl}). This process is repeated for the newly added nodes until a specified number of node hops is reached.

This technique for building a generic KG offers several advantages in comparison to other problem-specific approaches. First, it allows the same KG to be used for different variations of the task. It also enables transfer learning since KGs can be reused to tackle other related problems. Moreover, the construction of such a KG does not rely on expert knowledge. Besides, the structured representation of relationships between entities and concepts that KGs provide can be leveraged to generate robust embeddings for zero-shot learning.
% which is expensive and time-consuming.  
The trade-off is that such KGs are prone to noisy information in the used repositories. 

% In comparison, language models, such as BERT~\cite{devlin2018bert}, often rely on large amounts of unstructured text data to generate embeddings. While language models are highly effective at capturing semantic relationships between words and phrases, they can also be prone to create associations between concepts that are not actually related. This can lead to noisy or unreliable embeddings, which can in turn degrade the performance of zero-shot learning models. By contrast, the structured nature of KGs allows for more accurate and precise capture of relationships between entities and concepts, leading to more robust embeddings that can improve the accuracy and reliability of zero-shot learning models~\cite{brown2020language}.

\vspace*{0.0cm}\noindent\textbf{Computation of  Graph Embeddings (Stage 2)}:
We employ an established approach~\cite{Kampffmeyer2019, Wang2018b} that involves the training of a transformer-based Graph Convolutional Network (GCN)
 \textcolor{black}{ that utilizes a KG as input  %Training is performed %using features of a set of semantic entities acquired by a common sense repository, \textcolor{red}{(e.g. the ConceptNet, CSKG, or other)}  
 and generates an embedding vector for each node of the  KG. %. For the production of the embeddings vectors the GCM employs a sequence of transformations to the semantic features that correspond to the concepts linked to each node.
This process defines pre-computed GloVe word, i.e. semantic features~\cite{pennington2014glove}, for the KG nodes with each node representing a concept class.
% To compute node embeddings, the GNN is applied to encode the KG topology and the word feature embedding matrix. 
The GNN  aggregates each node's and its neighbors' features through a sequence of convolutions and pooling operations. %This results in the generation of a feature vector having a length equal to the dimension of the last layer in a visual CNN-based classifier that is instantiated using a ResNet-101 model. 
%By pre-training the visual classifier in a set of target classes 
The visual classifier is pre-trained on a set of target classes and using the weights of its fully connected layer, the GCN learns to produce visual feature representations, i.e. visual embeddings,  corresponding to the concept classes of the KG`s nodes.}
\textcolor{black}{
Formally,  the training involves the minimization of the L2 distance   $\mathcal{L_G}$ between the generated visual embeddings and the ground truth visual embeddings stemming from the visual classifier.} 
\textcolor{black}{In notation, 
\begin{equation}
\mathcal{L_G} = \frac{1}{2N} \sum_{n \in N} \sum_{p \in P} (W_{n,p} - \tildea{W}_{n,p})^2,
 \end{equation}
where $\tildea{W} \in \mathbb{R}^{|N|xP}$ denotes the weights of the GCN for the set of known concept classes $N$ and the dimensionality $P$ of the weight vector. Similar to~\cite{Kampffmeyer2019}, the ground truth weights, denoted as $W \in \mathbb{R}^{|N|xP}$, are obtained by extracting the last layer weights of a pre-trained CNN.}
% This process is repeated for all KG nodes corresponding to $S^U$ and $S^S$, generating semantic graph embeddings for all target state classes. 
%Each embedding comes in the form of a feature vector of length 2048. 

%By combining these embeddings for the $d$ target classes, a  $d \times 2048$ features matrix is defined that is integrated as the final layer of the visual CNN-based classifier that is employed in Stages~3-4.
%A critical aspect of this process is adjusting the GNN weights to align its predictions with the semantic space. This ensures that the semantic embeddings effectively aid the classifier used in Stages 3 and 4, operating in the visual space. 

\textcolor{black}{ 
The KG  given as an input to the GCN model is a hierarchical graph created for the requirements of the   ILSVRC 2012 dataset~\cite{russakovsky2015imagenet} and represents the WordNet hierarchical structure of the $1,000$ classes comprising the dataset. These 1,000 concept labels constitute the set of classes upon which the visual classifier used for the extraction of the ground truth visual embeddings is pre-trained.
}
% A critical aspect of this process is adjusting the GNN weights to align its predictions with the semantic space. This ensures that the semantic embeddings effectively aid the classifier used in Stages 3 and 4, operating in the visual space. 
% The concepts  that are used for the training refer to a set of 1K object classes of the ILSVRC 2012 dataset~\cite{russakovsky2015imagenet}, while the pre-trained ResNet101-based classifier is used for supervision to ensure that the GNN outputs, thus the semantic object class representations, are meaningfully embedded into the visual feature space. 
After the training is completed, the GCN model is employed to process the KG (constructed in Stage 1) and generate visual embeddings for the KG nodes that correspond to the object state classes,  by taking as input the  KG that was constructed during Stage 1. Each embedding comes in the form of a feature vector of length 2048, i.e. dimension of the last layer of the  pre-trained visual CNN-based classifier.
By combining these embeddings for the $d$ target classes, a  $d \times 2048$ features matrix is defined that is integrated as the final layer of the visual CNN-based classifier that is employed in Stages~3-4.

\vspace*{0.0cm}\noindent\textbf{Fine-tuning of the Visual Classifier (Stage 3)}:
The estimated semantic embeddings are integrated into a visual CNN classifier that relies on the ResNet backbone and is initially pre-trained for object classification. The embeddings serve as the final layer of the network, encapsulating the representations essential for predicting the train state classes $S^S$. To enable this adaptation, the visual classifier undergoes re-training, specifically tailored to the classification of the train classes. 
During this fine-tuning process, input images $I^T$ contain states sourced exclusively from the training set $S^S$, i.e. ``seen states''. The primary objective is to harness the classifier capabilities to classify these familiar states, accurately. Notably, fine-tuning involves keeping the weights of the last layer fixed, safeguarding the integrity of the acquired semantic representations from Stage 2. Consequently, adjustments are only applied to the weights of preceding layers to ensure they effectively match the ``frozen'' last-layer weights.
% Apart from this detail, the procedure takes place in the same manner as the training of a CNN classifier.
% in every training epoch a loss is computed the value of which guides the update of all layers weights except the last one. 
Following the notation introduced 
\textcolor{black}{in the beginning of Section~\ref{sec:method}, the loss function is defined as:}
\begin{equation}
% \mathcal{L} = -\sum_{i \in S^{S}} y_i \cdot \log(P(y=i|X))
\mathcal{L_V} = -\sum_{s \in S^S, i \in I^{T}} y_s \cdot \log(P(s|i)),
 \end{equation}
\textcolor{black}{for the predicted \textit{$y_s$} state label in the \textit{$S^S$} set of state labels. $P(s|i)$ denotes the probability of state label \textit{s} based on the softmax vector given an image \textit{i} from the $I^T$ training set.}

\noindent\textbf{Zero-shot OaSC (Stage 4)}:
Upon the completion of fine-tuning, the visual state classifier can be utilized for  prediction by choosing the most likely class
\begin{equation}
% \^y = \arg\max_{i \in S} \left( P(y=i|X) \right)
\hat{y} = \arg\max_{s \in S^U i \in I^{U}} \left( P(s|i) \right),
\end{equation}
\textcolor{black}{where $I^U$ denotes the test image set and $S^U$ the test state classes respectively.} 
We highlight that the classifier is well-suited for predicting either only unseen classes, i.e. zero-shot classification, or both seen and unseen classes, i.e. generalized zero-shot classification.
\vspace{-.15cm}
% \subsection{Pipeline}

% Overall, the pipeline of our method consists of four stages (\autoref{fig:pipeline}}). During \textbf{Stage 1}, the KG is constructed.

% \vspace*{0.2cm}\noindent\textbf{Construction of the KG (Stage 1)}:
% The KG creation process involves querying a common sense repository to enable generalization instead of creating a custom KG tailored to specific entities and relationships. Initially, nodes corresponding to the target state classes are generated. The repository is then queried for each node, and neighbors meeting specific criteria are added to the knowledge graph. This process continues for the newly added nodes until a specified number of hops is reached. More details can be found in the ablation section.

% \vspace*{0.2cm}\noindent\textbf{Computation of semantic embeddings (Stage 2)}:

% \vspace*{0.2cm}\noindent\textbf{Finetuning of the Classfier (Stage 3)}:

% \vspace*{0.2cm}\noindent\textbf{Deployment  (Stage 4)}:

\section{Experimental Evaluation}

\newcommand{
  \pbt}{\color{blue}{}
}

\begin{table*}[t]

    \centering
     % \resizebox{1\textwidth}{!}
      %  \resizebox{0.76\textwidth}{!}{\begin{minipage}{\textwidth} 
      \resizebox{2\columnwidth}{!}{
     \begin{threeparttable}
    \begin{tabular}{|l|c|c|c|c|c|c|c|c|c|c|c|c|c|c|c|c|}
   %  \begin{tabular}{|l|c|c|c|c|c|c|c|c|} 
   \hline
%\tiny
      
 % \diagbox[innerleftsep=.5cm,innerrightsep=0pt]{{\bf \multirow{2}{*}{Method}}}{{\bf \multirow{2}{*}{Dataset}}} 

\textbf{\centering  Method}  &\multicolumn{4}{|c|}{\textbf{OSDD}}  & \multicolumn{4}{|c|}{\textbf{CGQA-States}}  & \multicolumn{4}{|c|}{\textbf{MIT-States}}    & \multicolumn{4}{|c|}{\textbf{VAW-States}} \\      \cline{2-17} 

% \textbf{\centering  Method}  &\multicolumn{2}{|c}{\textbf{OSDD}~\cite{gouidis2022}}  & \multicolumn{2}{|c|}{\textbf{CGQA}~\cite{Mancini2022}}  & \multicolumn{2}{|c|}{\textbf{MIT}~\cite{Isola2015}}    & \multicolumn{2}{|c|}{\textbf{VAW}~\cite{Pham2021CVPR}} \\      \cline{2-9} 

 &\textbf{S} & \textbf{U}   
 & \textbf{HM} & \textbf{A} 
 &\textbf{S} & \textbf{U}  
 & \textbf{HM} & \textbf{A}
 &\textbf{S} & \textbf{U}  
& \textbf{HM} & \textbf{A}
 &\textbf{S} & \textbf{U}  
& \textbf{HM} & \textbf{A}
 \\ \hline \hline

\hline  

% AoP\cite{nagarajan2018attributes}   & 84.2 &  18.2  & 14.0 &  6.7
%  & 99.3 & 8.7 & 8.9 & 1.7
% & 100.0 &  55.5 &  37.5 &  30.6 \\ 

 AoP
\cite{nagarajan2018attributes}   
& 43.2 &  26.1  
& 20.7 &  7.4
 & 100.0 & 19.6 
 &22.9 & 13.3
& 100.0 &  11.6
&  13.2 &  7.0 
& 32.4 &  9.4 
&  9.5 &  2.0 \\ 

% LE+\cite{misra2017red} & 30.5 &     31.9  &  14.0 & 4.3
%      & 97.7 & 13.0 & 13.6 & 6.0
%  & 100.0 & 20.5 &   19.4 & 10.0 \\  

 LE+
\cite{misra2017red} 
& 30.5 &     31.9  
&  14.0 & 4.3
     & 97.7 & 12.5 
     & 12.8 & 5.5
 & 100.0 & 20.5
 &   19.4 & 10.0  
  & 56.5 & 16.8 
  &   15.9 & 5.8 \\  
%  TMN\cite{purushwalkam2019task} & 83.1 &  66.5  & 38.5 &  27.5
%  & 99.2 & 40.2 & 25.6 & 15.0
% & 100.0 &  17.7 & 17.8 &  11.0 \\   

  TMN
 \cite{purushwalkam2019task} 
 & 83.1 &  66.5  
 & \underline{38.5} &  \underline{27.5}
 & 99.2 & 40.2 
 & 25.6 & 15.0
& 100.0 &  17.8
& 17.7 & 11.0 
 & 86.7 &  55.2 
 & \bf 38.1 & \underline{27.1} \\  %cw
 % SymNet\cite{Li2020} &83.2 &     36.7  &  28.3 & 16.3 
 %     & 98.5 & 70.4  & 44.2 & 31.1  
 %     % & 49.3 & 41.7
 % & 94.1 & 22.3 &   29.6 & 17.3 \\  

  SymNet
  \cite{Li2020}
  &83.8 &     37.3  
  &  33.5 & 19.8
     & 99.2 & 24.5 
     & \underline{36.6} & \underline{20.6} 
     & 94.1 & 21.4 
     &   23.2 & 13.2
& 87.8 & 31.6 
&   37.3 & 21.5 \\  

%%.| Seen Acc attr: 78.91% | Unseen Acc attr: 49.73 | HM_attr: 32.10% | AUC_attr: 20.69% cw
%   Compcos\cite{Mancini2022}  & 86.0 &  60.0  & 36.0 &  25.4
%  & 99.2 & 22.8 & 21.9 & 10.1
% & 100.0 &  23.7 &  20.2 &  11.0 \\ 

  Compcos
 \cite{Mancini2022} 
 & 86.5 &  43.7  
 & 26.9 &  15.9 
 & 89.9& 17.1
 & 14.7 & 6.1
& 100.0 &  52.2
& \bf 36.4 &  \bf 25.8 
& 88.3 &  32.1
& 27.7 &  17.2 \\
%   KG-SP\cite{karthik2022open}  & 80.0 &  39.8  & 26.7 &  12.4
%  & 96.9 & 8.2 & 10.7 & 4.5
% & 100.0 &  6.5 &  12.2 &  6.3 \\ 
   KG-SP
  \cite{karthik2022open} 
  & 80.0 &  39.8  
  & 26.7 &  12.4
 & 96.9 & 8.2 
 & 10.7 & 4.5
& 100.0 &  7.1 
&  9.0 &  4.0 

& 83.9 &  11.4 
&  17.7 &  8.1 \\

  SCEN
\cite{li2022siamese} 
  & 77.8 &  41.5 
  & 35.2 &  22.5
 & 100.0 & 13.0 
 & 12.9 & 5.9
& 100.0 &  22.03
&  20.6 &  12.6 
& 89.6 &  37.4
&  28.2 &  17.3 \\

%   SCEN-NET\cite{li2022siamese}  & 49.2 &  38.3  & 19.8 &  7.1
%   & 100.0 & 64.9 & 46.4 & 39.5
% & 100.0 &  22.3 &  20.6 &  12.6 \\ 

%   IVR\cite{zhang2022learning}  & 85.8 &  38.5  & 36.0 &  23.1
%  & 98.5 & 18.8 & 17.1 & 8.4
% & 100.0 &  11.3 &  14.1 &  5.2 \\ 

   IVR
  \cite{zhang2022learning}  
  & 85.8 &  37.8  
  & 35.1 &  22.1
 & 98.4 & 18.8 
 & 17.1 & 8.4
& 100.0 &  11.3 
&  14.1 &  5.4  

& 88.9 &  11.0
&  16.2 &  7.8 \\
%   OADiS\cite{Saini_2022_CVPR}  & 80.9 &  65.8  & 26.1 &  16.5
%  & 96.1 & 11.7 & 11.1 & 5.6
% & 100.0 &  28.5 &  23.0 &  11.5 \\ 

   OADiS
  \cite{Saini_2022_CVPR} 
  & 72.7 &  55.5 
  & 23.1 &  13.0
 & 97.7 & 11.7
 & 11.9 & 4.8
& 94.1 &  30.0 
&  \underline{23.3} &  12.5 
& 83.3 &  53.5
&  33.8 &  23.9 \\ 
%   CANET\cite{wang2023learning}  & 85.6 &  36.4  & 20.2 &  12.1
%  & 100.0 & 9.5 & 11.3 & 5.1
% &  100.0 &  17.2 &  19.4 &  9.7 \\ 

%    CSP
%   \cite{csp2023} 
%  % & 85.6 &  36.4  
%   % & 14.8 &  6.1
%     & 28.0 &  15.0
%  %& 100.0 & 9.5 
%  & 37.8 &  28.6
%  %  45.03 &38.3
% %&  100.0 &  16.9 
% % &  44.48 &    39.48
% &   &    
% %&  87.8 &  53.4
% &  30.6 &  18.3 \\ 
% % &  37.7 &  26.6  \\ 
% %   ADE\cite{hao2023ade}  & 90.4 &  67.8  & 39.8 &  29.8
% %  & 100.0  & 68.5 & 44.9 & 38.0
% % & 100.0 &  23.7 &  28.3 &  17.2 \\ 

   CANET
  \cite{wang2023learning} 
  & 85.6 &  36.4  
  & 20.2 &  12.1
 & 100.0 & 9.5 
 & 11.3 & 5.0 
&  100.0 &  16.9 
&  23.1 &  11.9  
&  87.8 &  53.4
&  35.6 &  25.6  \\ 
%   ADE\cite{hao2023ade}  & 90.4 &  67.8  & 39.8 &  29.8
%  & 100.0  & 68.5 & 44.9 & 38.0
% & 100.0 &  23.7 &  28.3 &  17.2 \\ 
  
   ADE
  \cite{hao2023learning} 
  & 91.4 &  67.1 
  & \bf 40.5 &   \bf 30.3
 & 100.0  & 68.7
 &\bf 40.0 & \bf 33.3
& 100.0 &  24.9
&  22.6 & \underline{12.6}  
 % & 89.1 &  56.9 &  37.4 & 27.9   %cw

& 89.35 & 56.9 
& \underline{36.9} & \bf 27.6 \\ %ow
   
  \hline  \hline 
 
\centering  {\bf OaSC (ours)}
% (Ours)
  & 87.7 & 69.9 
  & \bf  \textcolor{red}{48.6} & \bf \textcolor{red} {39.8} 
   &97.1 & 73.4 
  & \bf  \textcolor{red}{43.6} & \bf  \textcolor{red}{36.5}
    &85.7 & 69.9 
  & \bf  \textcolor{red}{51.1} & \bf  \textcolor{red}{41.2}  
  
  &83.7 & 58.6 
  & \bf  \textcolor{red}{42.9} & \bf  \textcolor{red}{32.8}  

 \\  \hline 
% Improvement 
$\Delta$ \quad (gain) &{\cellcolor{gray}}  &{\cellcolor{gray}}  & +8.1 &  +9.5 & {\cellcolor{gray}} & {\cellcolor{gray}} &+3.6 &  +3.2 
 & {\cellcolor{gray}} & {\cellcolor{gray}} &+14.7 &  +15.4  & {\cellcolor{gray}} &{\cellcolor{gray}}  &+3.8 &  +5.2   
  
  \\  \hline  

%  \hline  
%   % CLIP  & 66.0 &  58.8  &  46.8 &   31.7
% % CLIP  & 57.5 &  69.8  &  29.5 &   22.3
% CLIP-ViTB16~\cite{radford2021learning} 
% %& 85.8 &  66.2 
% &  45.4 &   39.7
% %& 61.9 &  71.8 
% &  54.9 &   40.3
% %& 58.5 &  80.0 
% &  63.4 &   39.8

% %& 87.3 &  66.9 
% &  43.4 &   36.3 \\

% CLIP-ViTB32~\cite{radford2021learning} 
% %& 85.8 &  66.2 
% &  43.1 &   35.4
% %& 61.9 &  71.8 
% &  48.2 &   30.4
% %& 58.5 &  80.0 
% &  63.9 &   39.6

% %& 87.3 &  66.9 
% &  41.4&   33.4 \\

% CLIP-RN101~\cite{radford2021learning} 
% %& 85.8 &  66.2 
% &  33.6&   22.2 
% %& 61.9 &  71.8 
% &  51.8&   36.5
% %& 58.5 &  80.0 
% &  63.5 &   45.6

% %& 87.3 &  66.9 
% &  37.4 &   27.1 \\  \hline 
  \end{tabular}
 \caption{Aggregate results for the  Object Agnostic Setting. 
    Seen: Best Accuracy on seen classes. Unseen: Best accuracy on unseen classes. 
   HM: Best harmonic mean. A: Area under curve for the pairs of accuracy for seen and unseen classes.  Red/Bold/Underlined text indicates best/2nd best/3rd best performance.
   }
      \label{tab:table_aggr}
        \end{threeparttable}
       }
\end{table*}

%\vspace{-.35cm}
\subsection{Implementation and evaluation issues}
%\vspace{-.15cm}

\noindent\textbf{Implementation details}: 
The GNN was trained following the method outlined in Nayak et al. \cite{Kampffmeyer2019}. The model was trained for 1000 epochs on 950 randomly selected classes from the ILSVRC 2012 dataset \cite{russakovsky2015imagenet}, while the remaining 50 classes comprise the validation set. The model with the lowest validation loss was chosen to generate the seen and unseen class embeddings using the graph. For the seen classes, the embeddings were frozen, and a pre-trained ResNet101-backbone was fine-tuned on the individual datasets for 50 epochs using stochastic gradient descent  with a learning rate of 0.0001 and momentum of 0.9.

% For the training of the  GNN we follow the strategy propose in \cite{nayak:tmlr22} and train the model for 1000 epochs on 950 random classes
% from the ILSVRC 2012 \cite{russakovsky2015imagenet} while the remaining 50 classes are used for validation. The model with
% the least loss on the validation classes is used to generate the seen and unseen class embeddings with
% the graph.  We freeze the class embeddings for the seen classes and fine-tune a pretrained ResNet101-backbone on the individual datasets for 50 epochs using SGD with a learning rate 0.0001 and
% momentum of 0.9. 

% Currently with the exception of the OSDD dataset \cite{},  there are not exclusive object states dataset available, but rather attributes  dataset which include among their classes and object states. Therefore, we adopt two of the most popular attributes  datasets \cite{Isola2015} \cite{Mancini2022}  to our needs by taking the subsets that refer to object states in order to be used in the context of the experimental evaluation.

\vspace*{0.0cm}\noindent\textbf{Datasets}: 
Currently, there is a scarcity of datasets specifically designed for characterizing object states, 
% At present, there is a lack of datasets exclusively dedicated to object states, 
except for the OSDD~\cite{gouidis2022} which is a dataset tailored for state detection. Instead, existing attribute datasets include object states among their classes. To address this, we utilized two of the most widely used attribute datasets CGQA~\cite{Mancini2022} and MIT~\cite{Isola2015}, and extracted subsets that are specifically related to object states. We also experimented with VAW~\cite{Pham2021CVPR} which is a recently published object detection dataset that provides object state annotations for some of its samples. Regarding the OSDD and VAW, we extracted the bounding boxes of the original images to create images suitable for the OSC task. The complexity of each dataset can be assessed mainly by the number of unseen state classes and the average number of states per object class. More details on these datasets are presented in the supplementary section. 

\vspace*{0.0cm}\noindent\textbf{Metrics}: 
Our evaluation protocol follows the standard generalized zero-shot evaluation described in~\cite{Purushwalkam}, i.e., we calculate the Area Under the Curve (AUC) measuring the accuracy on both seen and unseen compositions at different operating points based on the bias term that is added to the scores of the unseen classes. The optimal zero-shot performance occurs when the bias term is positive, leading the classifier to prioritize the unseen labels. Conversely, the best seen performance is achieved with a negative bias term, which results in a focus on the seen labels. Additionally, we report the best harmonic mean (HM) which expresses a balance between the seen and unseen accuracy, respectively.

\noindent\textbf{Comparison with SOTA object-aware CZSL methods for state classification}: 
% Since there are no actual zero-shot state classifiers, we utilize SoA methods from the fields of  czsl and zero-shot attribute classification. In the case of czsl, they only method that can be used for ZS-OSC is  Compcos  which also exhibits the best performance in the czsl task. 
% As there are currently no pure zero-shot state classifiers available, we utilize  the state-of-the-art method from the field of CZSL\cite{Mancini2022} which involves both the prediction of object and state label for an image is  the most related field to to OSC.  We report the methods performance for three different settings: closed world, open world and object-oracle. For the first setting the method has to predict only among the valid object-state pairs, whereas for the second the method has to predict among all object-state pairs. For the third setting, all object labels are set to the generic term \'object\' enabling thus the method to predict only the state label. Although the closed world setting violates the zero-shot assumptions we include it as a baseline.
Given that there are currently no zero-shot state classifiers available, we resort to employing 11 state-of-the-art models \cite{misra2017red,nagarajan2018attributes,purushwalkam2019task,Li2020,Mancini2022,karthik2022open,li2022siamese,zhang2022learning,Saini_2022_CVPR,wang2023learning,hao2023learning}
from the field of Compositional Zero-Shot Learning (CZSL). These methods deal with predicting both object and state labels. As such, they are relevant to OSC - however, they are object-aware and not object-agnostic as the proposed OaSC method. 
% As these models are capable of producing state labels, they can be used in the context of OSC without any modifications.
We evaluate the performance of this approach on three different versions: %Object Agnostic, Closed World, and Open World:
\begin{itemize}[noitemsep,topsep=0pt]
\item {\bf Object Agnostic (OA)  version}: All object labels are replaced with the generic term ``object'', allowing the method to solely predict the state label.
\item {\bf Closed World (CW) version}: The method is tasked with predicting only among the valid object-state pairs. 
\item  {\bf Open World  (OW) version}: The method is tasked with predicting among all object-state pairs. 
\end{itemize}
In all three settings, we focus exclusively on the predictions concerning the states labels. It's important to emphasize that both the CW and OW versions of the models deviate from the principles of zero-shot conditions. Specifically, the CW version relies on pre-existing knowledge of valid states for each object, while the OW version considers a closed set of object labels corresponding to the states. These assumptions, although informative, limit the generality of the approach. Unlike these versions, our method remains entirely impartial to such constraints, demonstrating its versatility by maintaining consistency between training and inference. 

Additionally, it's noteworthy that both the CW and OW versions of the models incorporate knowledge about object categories, which is contrary to the object-agnostic assumption. In contrast, our approach remains consistent with the object-agnostic principle. Given these considerations, the fairest comparison to our method is the OA  version of the models. Nevertheless, for reference, we present the results of both the CW and OW versions of each model. This comprehensive approach provides a frame of reference while highlighting the distinct strengths of our method.

% It should be noted that both the CW and the OW versions violate the zero-shot conditions.  Namely, for the CW the valid states for each object are  known in advance, whereas, for the OW version, the set of object labels to which the states correspond is considered closed, which assumption sets it less generic than our approach, i.e. the same during training and inference, whereas our method is totally agnostic to this. In addition, both the CW and the OW  versions of the models use the knowledge for object categories, something that violates the object-agnostic assumption.  Therefore, the most fair comparison to our method is the GO version of the models. However, we report  the results of the  CW  and OW versions of each model as frames of reference. 
% Specifically, in the former case we can only use the Compcos method \cite{Mancini2022}, which has demonstrated the best performance in the context of czsl.

% The best zero-shot performance is achieved when the bias term is large, predicting only the unseen labels. The best seen performance is achieved when the bias term is negative, predicting only the seen labels.

% \begin{figure*}[t]
%     \centering
% \noindent\includegraphics[scale=0.6]{./figs/bar_chart_all.png}
% \caption{The ablation study for the GNN architecturesSupplementary.tex.}

%     \label{fig:ablation.svg}
% \end{figure*}

\subsection{Experimental results}

\noindent{\bf Intra-dataset evaluation:}
\autoref{tab:table_aggr} summarizes the results of the OA versions evaluation for the four employed datasets (the results for the CW and OW versions are presented as supplementary material due to space limitations). We report the performance of the version of
our model that was selected by the ablation study described in the next section. It is important to note that this version of the model does not
exhibit the best performance in all dataset categories.
Based on the obtained results, we observe that our method outperforms by a significant performance gain every other competing method. Specifically, in the MIT-States dataset OaSC outperforms by a margin of 15.4\% (14.7\% for HM)  the second best-performing method, which is the CompCos approach. Regarding the OSDD dataset, our  method outperforms the leading competitor, ADE,  with a gain of 9.5\% (8.1\% for HM). In the case of VAW, the gain in favor of our method is 5.2\% (3.8\% for HM) in comparison ADE which is the second-best method.  Lastly, in the CGQA-States dataset, our method demonstrates an improvement of 3.2\% (3.6\% for HM), surpassing the ADE model, which is the second best-performing among the competing methods in this scenario.
The substantial margin by which OaSC outperforms the competing methods in the OA setting indicates that the lack of information related to objects classes is detrimental for the CZSL methods. 
Moreover,  the fact that the CZSL methods in the OW and CW settings, although improved, are still inferior to  our method, suggests that the leveraging of KGs  can serve as a substitute for object-aware information.

\noindent{\bf Cross-dataset evaluation:} A further series of experiments was conducted concerning cross-dataset evaluation. \autoref{table:cross} reports the results obtained by our method and the ADE~\cite{hao2023learning}  model, which overall is the second-best model in the intra-dataset evaluation.  We can see that our method outperforms ADE in all cross settings when OSDD, CGQA and VAW  are used as training datasets, whereas ADE is better when MIT is used. %A plausible explanation for this is that MIT  is very distinct, visually, from the other 3 datasets and is also the smallest in terms of samples, which entails that the fine-tuning of a model in this dataset renders the classification in the other 3 datasets
%very challenging. 
This likely is due to   MIT  being very distinct, visually, from the other 3 datasets and being also the smallest in terms of samples, which entails that the fine-tuning of a model in this dataset renders the classification in the other 3 datasets
%very challenging. 
%Therefore, fine-tuning a model in this dataset renders the classification in the other  datasets very challenging. 
% The  backbone of ADE is a visual transformer (ViT) which is much more
% effective in learning representations than our CNN back-bone (ResNet101). %due to visual transformers having access to more sub-space global information across multi-head
% %attentions than CNNs . 
% Τhese more powerful  representations result in higher accuracy for  the  classes of the cross datasets in comparison to our CNN back-bone. 
ADE uses as a backbone a visual transformer (ViT) which is
much more effective in learning representations than our
CNN backbone (ResNet101), since visual transformers
have access to more sub-space global information across
multi-head attentions than CNNs. Therefore, 
the difference in backbones is crucial in this context.

%of the OA versions can be deemed only average, given the significantly smaller search space (i.e., set of states) in comparison to the search spaces of the Closed-CW (i.e., set of valid object-state pairs) and  Open-OW (i.e., set of all object-state pairs) worlds, respectively. 
% Overall, only the performance of the CW versions of some of the methods can be considered satisfactory.

 % The substantial margin by which our proposed method outperforms the object-based methods in every experiment supports the conclusion in the object-agnostic setting 
 % that the information of the object class is not a decisive factor in the task.  Additionally, by examining more closely the behavior of the three versions of the competing models (OA, OW, CW - see supplementary material) we can make the following observations. The OW versions perform in general very poorly, while the performance of the OA versions can be deemed only average, given the significantly smaller search space (i.e., set of states) in comparison to the search spaces of the Closed-CW (i.e., set of valid object-state pairs) and  Open-OW (i.e., set of all object-state pairs) worlds, respectively. 
% Overall, only the performance of the CW versions of some of the methods can be considered satisfactory.

\noindent{\bf Comparison with LPMs:}
We also report the performance of three variations of the CLIP~\cite{radford2021learning} model which is considered one of the best-performing Large Pre-Trained Models (LPMs) and is used extensively for a variety of downstream tasks such as state classification and BLIP~\cite{li2022blip} which also supports diverse downstream tasks but utilized mainly in the context of  Visual Question Answering. It is important to stress that although LPMs  are considered zero-shot learning models, they are rather classifiers in the wild since these models have been presented during their training with samples containing the target classes to which they are tested. However, since these models are witnessing wide popularity and are considered SoA methods, we opted to report these variants to serve as an additional frame of reference. The obtained results are summarized in~\autoref{table:clip}. We observe that OaSC performs better than CLIP-RN101 which is the CLIP variant that uses the same visual backbone as our classifier. In more detail, our method outperforms CLIP-RN101 by a margin of 17.7\% in OSDD  and by  5.1\% in the VAW, while it achieves the same performance in CGQA and falls short by  -4.4\% in MIT. Moreover, our model outscores  BLIP by  margins ranging from 10.4\% (CGQA) to 26.2 \%  (OSDD) across all datasets.  Overall,  these results provide a further indication of the power of our method. %This outcome suggests that LPMs tailored to the VQA task are not fitted to address the zero shot OaSC problem.
\begin{table}[t]

     \resizebox{\columnwidth}{!}{
     \centering
\begin{tabular}{|l|c|c|c|c|} 
   \hline

{\textbf{Variant}}  & {\textbf{OSDD}}~ \cite{gouidis2022}  & {\textbf{CGQA}}~\cite{Mancini2022}   & {\textbf{MIT}}~\cite{Isola2015}    &{\textbf{VAW}} \cite{Pham2021CVPR}  \\   

\hline\hline

{\bf RN101}
% %& 85.8 &  66.2 
% &  33.6
&   22.2 
% %& 61.9 &  71.8 
% &  51.8
&   36.5
% %& 58.5 &  80.0 
% &  63.5
&   45.6

% %& 87.3 &  66.9 
% &  37.4
&   27.1 \\  \hline 

{\bf ViT-B/16}%
% %& 85.8 &  66.2 
% &  45.4 
&   39.7
% %& 61.9 &  71.8 
% &  54.9 
&   40.3
% %& 58.5 &  80.0 
% &  63.4 
&   39.8

% %& 87.3 &  66.9 
% &  43.4 &   
&36.3 \\ \hline

{\bf ViT-B/32} %~\cite{radford2021learning} 
% %& 85.8 &  66.2 
% &  43.1 &   
&35.4
% %& 61.9 &  71.8 
% &  48.2 &  
&30.4
% %& 58.5 &  80.0 
% &  63.9 &   
&39.6

% %& 87.3 &  66.9 
% &  41.4&   
&33.4 \\
  
\hline
{\bf BLIP}  
& 13.6
&   26.1
&  27.2
  
& 16.1\\
\hline

  \end{tabular}
 } 
   \caption{AUC performance of CLIP for three different visual backbones. The models are fine-tuned as described in \cite{wortsman2021robust}. %The dataset splits are as in the experiments reported in \autoref{tab:table_aggr}.
   % The variant that is more close to our model is RN-101
   }
%\vspace{-0.2cm}
\label{table:clip}

\end{table}

\begin{table}[t]
\centering
%\tiny
%\begin{subtable}{0.48\linewidth}

     \resizebox{\columnwidth}{!}{
\begin{tabular}{|ll|llll|}
\hline
\multicolumn{2}{|l|}{{\bf Datasets}}           & \multicolumn{4}{c|}{{\bf OaSC (ours) vs ADE~\cite{hao2023learning} }}                                                         \\ \hline
\multicolumn{2}{|l|}{{\bf \diagbox[innerleftsep=.05cm,innerrightsep=4pt]{{\bf {Training}}}{{\bf {Testing}}}}} & \multicolumn{1}{c|}{\bf OSDD} & \multicolumn{1}{c|}{\bf CGQA} & \multicolumn{1}{c|}{\bf MIT} & \multicolumn{1}{c|}{\bf VAW} \\ \hline\hline
\multicolumn{2}{|l|}{{\bf OSDD}}       &\multicolumn{1}{|l|}{\cellcolor{gray}}  & 
 \multicolumn{1}{r|}{\textbf{24.8}/14.1}& 
 
\multicolumn{1}{r|}{\textbf{47.4}/27.3}       &  \multicolumn{1}{r|}{\textbf{16.9}/15.5}  \\ \hline
 \multicolumn{2}{|l|}{{\bf CGQA}}            & \multicolumn{1}{r|}{\textbf{35.0}/17.2}&\multicolumn{1}{r|}{\cellcolor{gray}}       &   \multicolumn{1}{r|}{\textbf{21.6}/12.8}        &  \multicolumn{1}{r|}{\textbf{34.4}/20.3}        \\ \hline
\multicolumn{2}{|l|}{{\bf MIT}}            &\multicolumn{1}{r|}{17.1/\textbf{19.1}} &\multicolumn{1}{r|}{6.1/\textbf{21.4}} &        \multicolumn{1}{r|}{\cellcolor{gray}}  &      \multicolumn{1}{r|}{10.1/\textbf{18.9}}   \\ \hline
\multicolumn{2}{|l|}{{\bf VAW}}     &          \multicolumn{1}{r|}{\textbf{23.0}/5.9}    & \multicolumn{1}{r|}{\textbf{29.3}/22.3}&\multicolumn{1}{r|}{\textbf{28.2}/3.1}    &          \multicolumn{1}{r|}{\cellcolor{gray}}   \\ \hline
\end{tabular}%
}
\caption{Cross-dataset evaluation of OaSC and ADE (AUC metric) for pairs of training (rows) and testing (columns) datasets.}

\label{table:cross}
% \end{subtable}
%\hspace{\fill}
%\vspace{-0.65cm}

\end{table}

\vspace{-.1cm}

\subsection{Ablation Study}\label{sec:abl}
\vspace{-.1cm}

% We ablate our method across the following categories: the GNN architecture, the KG source, the number of max hops that were used for the creation of the KG and the policy that were followed w.r.t. the inclusion of nodes to the KG. For each ablated model we report the best accuracy achieved on seen and unseen classes, the best harmonic mean and the best AUC for each of the three datasets. The results are presented in \autoref{tab:ablation}.

We conducted a host of ablation experiments across several problem dimensions to select the optimal parameters for our model.  Specifically, we explored the impact of varying the GNN architecture, the KG source, the maximum number of hops used for KG creation and the policy for including nodes in the KG. Due to space limitations, the performance exhibited by every ablated model is provided in the supplementary material. Here, we present aggregated means of all models across each of the ablated dimensions reporting the best harmonic mean and the AUC for each of the four datasets, respectively. 
% The results of these experiments are presented in \autoref{tab:abl1} - \autoref{tab:abl3}.

%%%AAA: Parakatw, prwta parousiazeis oles tis diastaseis tou ablation, kai meta ola ta relevant results. Gia na veltiwseis to locality of reference kanto alliws: ablation diastasi 1 - results 1, ablation diastasi 2 - results 2...

\vspace*{0.0cm}\noindent\textbf{GNN architecture}:
We conduct experiments using 4 different GNN architectures:  GCN \cite{kipf2016semi}, R-GCN \cite{schlichtkrull2018modeling},   LSTM \cite{hamilton2017inductive} and Tr-GCN \cite{nayak:tmlr22}. The ablation results for the different architectures are presented in \autoref{tab:abl1}. 
The Tr-GCN framework outperforms the other frameworks in all datasets w.r.t. AUC metric, whereas it scores best w.r.t. HM metric in the OSSD and VAW and comes second in the two other datasets. The R-GCN framework exhibits the second-best performance, while the GCN framework comes in third and the LSTM framework exhibits the worst performance. 
% These findings are consistent with prior research in the domain of zero-shot object classification and validate the view that the architecture of .

\vspace*{0.0cm}\noindent\textbf{KG source}:
% The KG sources that we use are: ConceptNet\cite{speer2017conceptnet} and WordNet\cite{fellbaum2010wordnet}. We also include in our experiments some KGs that were created using information from both sources. We mention also that we attempted to utilize other sources such as Dbpedia\cite{} and WikiData\cite{} but did not succeed at finding the necessary information for the creation of a KG.
We employed two KG sources, namely ConceptNet~\cite{speer2017conceptnet} and WordNet~\cite{fellbaum2010wordnet}, and also experimented with combining information from both sources. Other sources such as Dbpedia~\cite{auer2007dbpedia} and WikiData~\cite{vrandevcic2014wikidata} were also considered, but the necessary information for constructing a KG could not be obtained. To better assess the contribution of the KGs, we include a ConceptNet-based model in which the target state classes were mapped to other unrelated state embeddings of the KG and a random model where the embeddings corresponding to the target state classes were generated by a random process. 

Based on the results reported in \autoref{tab:abl2}, the ConceptNet-based model outperforms WordNet across all four datasets, while combining both sources results in performance gains for the HM metric across all four datasets and for the AUC metric in three of the datasets. The difference in favor of ConceptNet can be attributed to the difference between the type of information that each KG holds. ConceptNet contains mainly common-sense knowledge and also includes some lexicographic information, while WordNet contains only lexicographic information. Still, the fact that the best results are achieved by a model that uses both sources suggests their complementarity.  
% Taken together, these findings offer substantial support for \textbf{Hypothesis 1}.  

\begin{table}[t]
%	\small
    \centering
  
 %\begin{subtable}{0.48\linewidth}

   \resizebox{\columnwidth}{!}{
    \begin{tabular}{|l|c|c|c|c|}

\hline
 \bf \diagbox[innerleftsep=.05cm,innerrightsep=4pt]{{\bf {Dataset}}}{{\bf {Arch}}} & 
  \textbf{LSTM}& \textbf{GCN}   & \textbf{R-GCN}& \textbf{Tr-GCN}\\
 
\cline{2-5}
\hline
\hline
\textbf{OSDD}  & 39.0 / 25.7 & 40.0 / 27.0 & 42.9 / 29.9 & \bf 43.2 / 30.3  \\
\hline  
\textbf{CGQA}  & 28.3 / 37.8 & 30.6 / 40.2 &  \textbf{29.0} / 38.1 &28.2 / \textbf{38.5} 
  \\ \hline

\textbf{MIT} & 47.7 / 30.7 & 50.7 / 34.3  & \textbf{53.7} / 36.6 & 51.2 / \textbf{39.8}  \\   \hline  
\textbf{VAW} & 32.2 / 22.1 & 34.3 / 23.4  & 36.5 / 25.6 & \textbf{39.2} / \textbf{27.6}  \\   \hline  
\end{tabular}
%\end{minipage}
}
\caption{Ablation results for the framework architecture.  The first (second) value in each cell corresponds to the best HM (AUC). } 

% All values are aggregate averages.}
\label{tab:abl1}
 %\vspace{-0.4cm} 

% \end{subtable}
\end{table}

\begin{table}[t]
	%\tiny
     \centering
  %     \resizebox{1\columnwidth}{!}{
%\begin{subtable}[t]{0.48\linewidth}

     \resizebox{\columnwidth}{!}{
    \begin{tabular}{|l|c|c|c|c|c|}

\hline
 \bf \diagbox[innerleftsep=.05cm,innerrightsep=4pt]{{\bf {Dataset}}}{{\bf {KG }}} & 
  \textbf{CN}& \textbf{WN}   & \textbf{CN+WN}&  \textbf{IE}&   \textbf{RN} \\

\hline\hline

\textbf{OSDD}  &43.5/30.5 & 32.6/18.5 &  \bf 45.4/34.7  & 19.4/9.3   & 8.2/3.1\\ \hline

\textbf{CGQA}  &  39.2/29.1  & 37.9/27.4 & \bf 44.5/34.7 &  20.1/9.0  & 11.1/5.7 
  \\ \hline

\textbf{MIT} & 53.3/\textbf{42.6} & 38.5/26.6  & \textbf{54.0}/42.1    & 33.8/22.1 & 18.6/13.0 \\   \hline  
\textbf{VAW} & 41.0/28.1 & 31.0/17.3  & \textbf{39.2}/\textbf{32.1}    & 15.3/19.2 & 7.3/3.5 \\   \hline
% \textbf{AGR} & 43.5/30.5 & 32.6/18.5 &  45.4/34.7 & 39.2/29.1  & 37.9/27.4 & 44.5/34.7 & 53.3/42.6 & 38.5/26.6  & 54.0/42.1 \\     

\end{tabular}
}
  \caption{Ablation results for the KG source. The first (second) value in each cell corresponds to the best HM (AUC). CN: ConceptNet. WN: WordNet, WN+CN: Model based on both ConceptNet and WordNet. IE: ConceptNet-Based Model (irrelevant embeddings).  RN:  Model with random embeddings. }

% The values in the three fist columns are aggregate averages.}
\label{tab:abl2}
%\end{subtable}
 % \vspace{-0.5cm} 
 \end{table}

Furthermore, the performance of the model using the random embeddings is very low, whereas the  ConceptNet-based model using unrelated state embeddings achieves a clearly better performance which remains significantly lower than that of the other CN-based models. 
The distinction between these approaches can be attributed to the
distribution of their embeddings: the former model employs a balanced and representative distribution enabled by GNN which permits the model to map the learned representations to the visual information of seen classes during the fine-tuning procedure. In contrast, the latter
model has a completely random distribution that cannot be mapped to the semantic representations. The unrelated embeddings do not leverage the recognition of unseen classes, thus resulting in the lower performance of the model. This is further supported by the results included in the supplementary material where the best seen and unseen accuracies are also
reported.  
% The distinction between these approached lies in their
% distribution: the former model employs a balanced and rep-
% resentative distribution enabled by GNN, while the latter
% model has a completely random distribution. This suggests
% that the fine-tuning process can yield competitive seen accuracy even with unrelated embeddings to the target labels as long as the distribution is appropriate. In contrast, achieving accuracy for unseen classes requires an exact mapping
% between the embeddings and the target states

\begin{table}[t]
%	\small
     \centering
    %   \resizebox{1\columnwidth}{!}{
%\begin{subtable}[t]{0.48\linewidth}

         \resizebox{\columnwidth}{!}{
    \begin{tabular}{|l|c|c||c|c|}

\hline
 \bf \diagbox[innerleftsep=.05cm,innerrightsep=3pt]{{\bf {Dataset}}}{{\bf {Hops/Policy}}} & 
  \textbf{Hop 2}& \textbf{Hop 3}   &  \textbf{NP}& \textbf{THR}
 
\\
\hline\hline

\textbf{OSDD}  & \bf 43.1/30.6  &   41.0/27.6 & 38.8/25.3   & \bf 42.5/28.5 \\
\hline
\textbf{CGQA-States}  &30.3/39.5 & \bf 31.4/41.0 &  25.9/36.0  &   \bf 29.8/39.5
  \\ \hline

\textbf{MIT-States} & 52.3/\textbf{36.9} &  \textbf{54.8}/36.5 & 45.9/31.7 & \bf 56.0/42.3 \\   \hline  
\textbf{VAW} & \textbf{37.5}/\textbf{27.8} &  35.5/24.3 & 31.5/20.7 &  \bf 34.1/\bf 23.0  \\   \hline

% \textbf{Aggr} &42.3/29.6  &  43.2/30.4 & 39.6/26.6  & 41.0/27.6 &
% 30.1/39.8  & 28.4/37.7 & 22.1/32.3  &  31.4/41.0 &
% 51.7/35.5 & 52.5/36.8 & 44.0/32.1 & 54.8/36.5 \\

\end{tabular}
}
    \caption{Ablation results. 
    1st (2nd) column, number of hops: average performance of models that are based on a KG with a number of hops equal to 2 (3).
    3rd (4th) column, threshlold policy:  average performance of models that are based on a KG created without (with) threshold policy. The 1st (2nd) value in each cell corresponds to the best HM (AUC).}
    
%}

% All values are aggregate averages.}
\label{tab:abl3}
%\end{subtable}
\vspace{-0.3cm} 
%\vspace{-0.65cm}

\end{table}

\vspace*{0.0cm}\noindent\textbf{Number of max node hops}:
We experiment with a hop count equal to 2 and to 3 for both KGs. The results are shown in \autoref{tab:abl3}. No consistent pattern can be identified.  The best average performance is achieved for the OSDD and VAW datasets at hop 2, while the best average performance is exhibited for the CGQA-State dataset at hop 3. In MIT-States there is no clear winner, as hop 2 shows superior AUC and hop 3 exhibits superior HM. This suggests that introducing nodes beyond a certain limit may introduce noise and potentially deteriorate the overall performance in specific cases, as observed in the OSDD dataset. 
% This outcome is consistent with the \textbf{Hypothesis 1}.

\vspace*{0.0cm}\noindent\textbf{Node policy}:
We investigate two strategies for adding nodes to our knowledge graph, indiscriminate inclusion of all neighboring nodes and selective inclusion of only relevant nodes. To determine relevance in ConceptNet, we use the edge weight between the queried node and its neighbors as the inclusion criterion. In WordNet, we use the Wu-Palmer Similarity metric~\cite{wu1994verb} 
between the two nodes. Additionally, in WordNet, we explore a hierarchical policy of accepting candidate nodes only if their ancestors belong to certain generic categories, such as attributes or objects. 
The results (last two columns of \autoref{tab:abl3}) show that adopting this policy leads to significant performance improvements across all three datasets. This finding complements the previous observation regarding the number of hops and further strengthens the notion that the presence of noisy nodes can have a detrimental effect on model performance. 
\section{Summary}

This work introduced OaSC, a novel method for zero-shot object-agnostic state classification.  OaSC leverages knowledge graphs and graph neural networks to infer object states without relying on object class information, enabling it to generalize to unseen objects.  Our extensive evaluation on four benchmark datasets demonstrated OaSC superior performance compared to SOTA CZSL methods. Furthermore,   the extensive comprehensive ablation study provided valuable insights into the impact of different design choices on the method's performance. %, highlighting the importance of KG construction and GNN architectures. 

\vspace*{0.0cm}\noindent\textbf{Acknowledgements} The Hellenic Foundation for Research and Innovation (H.F.R.I.) funded this research project under the 3rd Call for  H.F.R.I. Research Projects to support Post-Doctoral Researchers (Project Number 7678 InterLinK: Visual Recognition and Anticipation of Human-Object Interactions using Deep Learning, Knowledge Graphs and Reasoning) and under the “1st Call for H.F.R.I Research Projects to support Faculty members and Researchers and the procurement of high-cost research equipment”, project I.C.Humans, (Project Number  91).

%\vspace*{0.0cm}\noindent\textbf{Acknowledgements} The Hellenic Foundation for Research and Innovation (H.F.R.I.) funded this research under the Projects 1) InterLinK (n. 7678) and 2) I.C.Humans (n. 91).
 %\newpage
{\small
 \bibliographystyle{splncs04}
\bibliography{egbib}
}
\clearpage
\newpage
% \appendix

%\section{Supplementary Section}

% We present material that due to space limitations could not be included in the main paper. In Section~\ref{exp} we present and discuss the results for the CW and OW versions evaluation (section 4.2 main paper), whereas in Section~\ref{abl} we present and discuss the full results of the ablation study (section 4.3 of main paper). 

\section*{Supplementary Material}

\subsection{Datasets Details}
\autoref{tab:datasets} 
presents the following details for each dataset: i) the number of the training, validation and test samples; ii) the number of state and object classes; iii) the valid and  iv) the total object-state combinations and v) the average
  number of states in which an object can be situated.
\subsection{Evaluation of the CW and OW versions}

\label{exp}
The results for the Open World (OW) and Closed World (CW) versions of the models are shown in \autoref{tab:table_aggr_open} and  \autoref{tab:table_aggr_closed},   respectively.
For the OW settings our method continues to outperform the competing methods, although the performance gain has predictably been decreased. Moreover, w.r.t  OSDD dataset, the 2nd  best method is  IVR~\cite{zhang2022learning}, whereas CANET~\cite{wang2023learning} is the 3rd best method. In the case of the CGQA-States dataset, the 2nd and 3rd best method is  IVR~\cite{zhang2022learning}  and   CANET~\cite{wang2023learning}, respectively. Concerning the MIT-States dataset the 2nd  best method is the IVR\cite{zhang2022learning}, whereas KG-SP~\cite{karthik2022open} exhibits the 3rd best AUC score and   CANET~\cite{wang2023learning} the 3rd best HM score. Finally, in the case of the VAW dataset, the 2nd  best performance  is achieved by  CANET~\cite{wang2023learning}, while IVR~\cite{zhang2022learning} ranks 3rd.

Regarding the CW settings, our method ranks 1st for the OSDD, VAW and MIT-states datasets and 4th for the  CGQA-states dataset. Regarding the OSDD dataset,  IVR~\cite{zhang2022learning} exhibits the 2nd best performance and  KG-SP~\cite {karthik2022open} the 3rd best performance. In the case of MIT-States dataset,   CompCos~\cite{Mancini2022}  achieves the 2nd best performance and   ADE~\cite{hao2023learning}  the 3rd best performance. Concerning the CGQA-states dataset, the best performance is achieved by   CANET~\cite{wang2023learning}, the 2nd best by   CompCos~\cite{Mancini2022}  and the 3rd best by  OADiS\cite{Saini_2022_CVPR}. Finally, regarding VAW, the 2nd best method is ADE~\cite{hao2023learning} and the 3rd best method is CANET~\cite{wang2023learning}.

\subsection{Additional Results of the Ablation Study}\label{abl}
\autoref{tab:KGS} outlines the details of the employed KGs, while \autoref{tab:ablation} summarizes the performance of all ablated models across the four datasets.

\noindent 1st Sub-table (GNN Architectures): The Tr-GCN-based model CN+WN\_H2\_TH\_GCN demonstrates the best overall performance.

\noindent 2nd Sub-table (KGs): The ConceptNet-based model CN\_H2\_TH\_Tr-GCN achieves the highest scores.

\noindent 3rd Sub-table (Hops): Most models achieve their best performance with two hops.

\noindent 4th Sub-table (Node Policy): Adopting a node policy slightly improves the performance of most models.

Notably, while CN\_H2\_TH\_Tr-GCN achieves the best scores on two of the three datasets, CN+WN\_H2\_TH\_GCN was selected for comparison with competing methods, as this selection was based on aggregate averages across all four categories.

In seen classes, the model using unrelated embeddings (CN\_H3\_UN\_Tr-GCN) achieves similar accuracy to its counterpart with standard embeddings (CN\_H3\_Tr-GCN). However, CN\_H3\_UN\_Tr-GCN performs significantly worse in unseen classes, with its HM and AUC scores being three to four times lower than those of CN\_H3\_Tr-GCN. In contrast, the random model performs poorly across all metrics.

The key distinction between CN\_H3\_UN\_Tr-GCN and the random model lies in their embedding distributions: in the former, the GNN enables a balanced and representative distribution, while in the latter, the distribution is entirely random. This suggests that fine-tuning can yield competitive accuracy for seen classes even when embeddings are unrelated to target labels, provided they are distributed effectively. However, for unseen classes, accuracy depends on a precise mapping between embeddings and target labels.

\begin{table*}[t]
    \centering
%\resizebox{\textwidth}{!}{ 
    \begin{tabular}{|l|rrr|rrrrr|}

\hline 
\textbf{Dataset} & \textbf{Train}  &  \textbf{Val}   &  \textbf{Test} & \textbf{States} & \textbf{Objects} & \textbf{VOSC} & \textbf{TOSC} & \textbf{S\textbackslash O} \\ \hline \hline

OSDD \cite{gouidis2022} &   6,977 & 1,124  & 5,275 & 9 & 14 & 35 &126 &2.36 \\ \hline 
CGQA-states \cite{Mancini2022} &   244 & 46 & 806 & 5 & 17  &41 & 75 & 1.71  \\ \hline
MIT-states \cite{Isola2015} &  170 & 34 & 274 & 5 & 14 & 20 &  70  & 1.57 \\\hline
VAW \cite{Pham2021CVPR} &  2,752 & 516 & 1,584 & 9 & 23 & 51 &  207  & 2.61 \\

   \hline
  \end{tabular}
%}   
   \caption{Details about the four image datasets utilized in this work. Train/Val/Test: Number of Training/Validation/Testing Images. States:  Number of State classes, Objects: Number of Object classes. VOSC/TOSC: Valid/Total Object-State combinations. S\textbackslash O: Average number of states than an Object can be situated in.}
    \label{tab:datasets}
\end{table*}

% Additionally, it can be observed that the model employing irrelevant embeddings (CN\_H3\_UN\_Tr-GCN) performs similarly to the model using standard embeddings (CN\_H3\_Tr-GCN) in terms of accuracy for seen classes. However, the accuracy for unseen classes, as well as HM and AUC scores, are three to four times lower in CN\_H3\_UN\_Tr-GCN compared to CN\_H3\_Tr-GCN. On the other hand, the random model exhibits poor performance in all four metrics. 
% The distinction between the embeddings of CN\_H3\_UN\_Tr-GCN and the random model lies in their distribution: the former model employs a balanced and representative distribution enabled by GNN, while the latter model has a completely random distribution. This suggests that the fine-tuning process can yield competitive seen accuracy even with unrelated embeddings to the target labels as long as the distribution is appropriate. In contrast, achieving accuracy for unseen classes requires an exact mapping between the embeddings and the target states. 

\begin{table*}[t]
	%\tiny
   
 \resizebox{0.9\textwidth}{!}{\begin{minipage}{\textwidth} 

     \begin{threeparttable}
      \centering
    \begin{tabular}{|l|c|c|c|c|c|c|c|c|c|c|c|c|c|c|c|c|c|c|}
     
   \hline

\textbf{\centering  Method}  &\multicolumn{4}{|c|}{\textbf{OSDD}}  & \multicolumn{4}{|c|}{\textbf{CGQA-States}}  & \multicolumn{4}{|c|}{\textbf{MIT-States}}  & \multicolumn{4}{|c|}{\textbf{VAW}} \\      \cline{2-17} 

 &\textbf{S} & \textbf{Un}   & \textbf{HM} & \textbf{AUC} & \textbf{S} & \textbf{Un}   & \textbf{HM} & \textbf{AUC}
& \textbf{S} &\textbf{Un}  & \textbf{HM} & \textbf{AUC}
& \textbf{S} &\textbf{Un}  & \textbf{HM} & \textbf{AUC}
 \\ \hline \hline

\hline  

AoP~\cite{nagarajan2018attributes}   &69.9 &  33.3  & 31.6 & 13.3
 & 14.5 & 4.3 & 4.4 & 0.3
& 36.4 &  4.8 &  8.4 &  1.3

& 59.6 &  5.4 &  6.1 &  1.3\\

LE+~\cite{misra2017red} & 71.6 &     14.3  &  20.8 & 6.5
     & 29.1 & 4.0 & 7.0 & 0.6
 & 45.5 & 14.9 &   15.1 & 4.3    
  & 23.7 & 12.3 &   13.7 & 0.4 \\  
 TMN~\cite{purushwalkam2019task} & 73.4 &  43.6  & 33.7 &  19.0
 & 45.5 & 29.7 & 19.3 & 6.1
& 69.7&  18.4 & 22.4 &  6.3   
& 77.6&  35.5 & 26.8 &  14.3 \\

 SymNet~\cite{Li2020} &77.7 &  14.0 & 21.1 & 7.5 
     & 94.0 & 7.1 & 13.7 & 6.1
 & 97.0 & 1.9 &   2.1 & 0.9 
& 82.2 &  3.1 &  3.5 &  1.2\\

  CompCos~\cite{Mancini2022}  &  78.7 &  31.5  & 42.0 & 22.1
   & 95.5 & 4.0 & 7.7 & 3.4  
   &   75.8 & 2.5 &   4.9&  1.2 
   &   75.8 & 2.5 &   4.9&  1.2 \\
  KG-SP\cite{karthik2022open}  & 77.0 &  29.8 & 35.4 &  17.9
 % & 95.5 & 2.6 & 1.1 & 0.5
  & 94.0 & 16.9 & 26.1 & 12.7
% & 97.0 &  1.2 &  1.9 &  0.4 \\ 
& 97.0 &  15.5 & 22.6 & 12.0
& 74.3 &  12.3 & 17.6 & 8.6
 \\ 

  SCEN-NET~\cite{li2022siamese}  & 75.8 &  25.5  & 26.3 &  10.7
 & 83.6 & 7.4 & 13.6 & 5.9
& 36.4 &  8.5 &  13.0 &  1.6  
& 22.0 &  12.0 &  11.1 &  2.5 \\ 

  IVR~\cite{zhang2022learning}  & 78.8 &  61.6  & \bf 44.2 &  \bf 30.8
 & 94.0 & 40.3 & \bf {37.4} & \bf {26.4}
& 96.9 &  22.5 &  \textbf{24.5} & \bf 14.9 
& 87.2 &  37.4 &  \underline{29.7} &  \underline{18.2} \\ 

  OADiS~\cite{Saini_2022_CVPR}  & 76.5 &  20.5  & 27.1 &  10.7
   % & 96.1 & 11.7 & 11.1 & 5.6
 & 94.8 & 26.3 & 20.3 & 12.0
& 93.9 &  29.1 &  23.4 &  \underline{12.5} 
& 82.8 &  8.9 &  11.0 &  4.2 \\ 

% & 100.0 &  28.5 &  23.0 &  11.5 \\ 
  CANET~\cite{wang2023learning}  & 79.2 &  43.9  & \underline{43.7} &  \underline{27.2}
 & 95.5 & 51.3 &  \underline{41.9} &  \underline{26.1}
&  96.9 &  19.3 &  \underline{22.7} &  11.4 

&  90.1 &  53.9 &   \bf  40.4 &  \bf 29.7 \\ 
  ADE~\cite{hao2023learning}  & 80.2 &  27.6 & 32.3 &  12.3
 & 95.5  & 16.3 & 25.7 & 12.8
& 78.8 &  4.5 &  4.7 &  0.8 
  
& 80.8&  22.3 &  14.3&  8.4 \\ 
  
    \hline  \hline

\centering  OaSC (Ours)
  & 87.7 & 69.9 & \bf  \textcolor{red}{48.6} & \bf \textcolor{red} {39.8} & 
  97.1 & 73.4 & \bf  \textcolor{red}{43.6} & \bf  \textcolor{red}{36.5} & 
  85.7 & 69.9 & \bf  \textcolor{red}{51.1} & \bf  \textcolor{red}{41.2} 
&83.7 & 58.6   & \bf  \textcolor{red}{42.9} & \bf  \textcolor{red}{32.8}  \\  \hline

  \end{tabular}

   \caption{   \label{tab:table_aggr_open}Aggregate results for Open World Versions. S: Best Accuracy on seen classes. UN: Best accuracy on unseen classes. HM: Best harmonic mean. AUC: Area under curve for the  pairs of accuracy for seen and unseen classes. Red/Bold/Underlined text indicates best/2nd best/3rd best performance.
   }
        \end{threeparttable}
       \end{minipage} }
   \end{table*}

\begin{table*}[h!]

  \resizebox{0.9\textwidth}{!}{\begin{minipage}{\textwidth} 
     \begin{threeparttable}
      \centering
    \begin{tabular}{|l|c|c|c|c|c|c|c|c|c|c|c|c|c|c|c|c|c|c|}
    
   \hline

\textbf{\centering  Method}  &\multicolumn{4}{|c|}{\textbf{OSDD}}  & \multicolumn{4}{|c|}{\textbf{CGQA-States}}  & \multicolumn{4}{|c|}{\textbf{MIT-States}}  & \multicolumn{4}{|c|}{\textbf{VAW}} \\      \cline{2-17} 

 &\textbf{S} & \textbf{UN}   & \textbf{HM} & \textbf{AUC} & \textbf{S} & \textbf{UN}   & \textbf{HM} & \textbf{AUC}
& \textbf{S} &\textbf{UN}  & \textbf{HM} & \textbf{AUC}
& \textbf{S} &\textbf{UN}  & \textbf{HM} & \textbf{AUC}
 \\ \hline \hline

AoP~\cite{nagarajan2018attributes}   &75.9 &  53.5  & 32.2& 19.5
 & 95.5 & 50.0 & 35.9 & 27.8
& 48.5 &  20.9 &  15.1 &  4.1
& 55.1 &  44.7 &  24.1 &  11.6\\

LE+~\cite{misra2017red} & 68.6 &     31.7  &  34.5 & 16.9
     & 93.5 & 16.1 & 16.1 & 8.1
 & 63.6 & 14.6 &   20.3 & 7.1   
  & 41.6 & 2.3 &   2.6 & 1.2  \\ 
 TMN\cite{purushwalkam2019task} & 71.5 &  49.8  & 35.0 &  20.8
 & 97.0 & 76.0 & 39.9 & 32.2
& 84.9 &  30.7 & 27.4 &  16.1    
& 82.6 &  55.5 & 37.3 &  25.6 \\

 SymNet~\cite{Li2020} &77.7 &  59.4 & 44.2 & \underline {31.0} 
     & 95.5 & 27.4 & 39.4 & 24.4
 & 96.9 & 27.5 &   26.8 & 15.7  
  
 & 89.2 & 46.6 &   40.0 & 27.4 \\

  Compcos~\cite{Mancini2022}  &  76.3 &  45.3  & 38.7 & 23.8
    & 92.5 & 73.9 & \bf 48.1 & \bf 41.5  

   &   100.0 & 44.9 &  \textbf{32.3} &   \textbf{23.8} 
   &   88.4 & 51.4 &  39.3 &   29.1 \\
  KG-SP~\cite{karthik2022open}  & 78.0 &  55.0  & \bf  {47.6} & 29.7
 & 95.5 & 17.7 & 27.2 & 13.5
& 97.1 &  15.5 &  22.6 &  12.0  
& 89.4 &  37.3& 39.3 & 23.4\\ 
  SCEN-NET\cite{li2022siamese}  & 75.1 &  45.6  & 39.4 &  22.7
 & 94.1 & 53.4 & 41.1 & 31.0
& 84.9 &  23.1 &  22.1 &  11.5 
& 90.5&  44.2 &  37.7&  23.5 \\ 

  IVR~\cite{zhang2022learning}  & 78.4 &  60.5  & \underline{46.0} &  \bf {31.8} 
 & 94.0 & 43.4 & 35.2 & 25.2
& 87.9 &  28.8 &  27.1 &  14.0  
& 86.7 &  38.2 &  30.5 &  18.5 \\ 

%   OADiS\cite{Saini_2022_CVPR}  & 78.7 &  59.7  & 38.3 &  26.2
%  & 96.1 & 11.7 & 11.1 & 5.6
% & 100.0 &  28.5 &  23.0 &  11.5 \\ 

  OADiS~\cite{Saini_2022_CVPR}  & 78.7 &  59.7  & 38.3 &  26.2
 & 95.5 & 78.6 & 43.5 & \underline{36.7}
& 93.9 &  29.4 &  28.3 &  17.2 
& 89.9 & 61.8 &  39.8&  \underline{30.5}\\ 

  CANET~\cite{wang2023learning}  & 80.3 &  43.6  & 45.1 &  27.9
 & 95.5 & 64.9 & \bf  \textcolor{red}{50.0} & \bf  \textcolor{red}{43.3}
&  96.9 &  23.0 &  28.2 &  15.9  
&  90.3 &  54.6 &  \underline{40.8} & \underline{30.5} \\ 
  ADE\cite{hao2023learning}  & 82.0 &  42.5  & 35.9  &  20.6
 & 94.8  & 58.3 & \underline{45.5} & 34.9
& 93.9 &  27.5 &  \underline{30.4} & \underline{19.2} 
& 90.7&  45.0 & \bf 40.9& \bf 30.6 \\

    \hline  \hline 
 
\centering  OaSC (Ours)
  & 87.7 & 69.9 & \bf  \textcolor{red}{48.6} & \bf \textcolor{red} {39.8} & 
  97.1 & 73.4 &   43.6 &  36.5 & 
  85.7 & 69.9 & \bf  \textcolor{red}{51.1} & \bf  \textcolor{red}{41.2} 
&83.7 & 58.6   & \bf  \textcolor{red}{42.9} & \bf  \textcolor{red}{32.8}
\\  \hline

  \end{tabular}

   \caption{   \label{tab:table_aggr_closed}Aggregate results for Closed World Versions. S: Best Accuracy on seen classes. UN: Best accuracy on unseen classes. HM: Best harmonic mean. AUC: Area under curve for the  pairs of accuracy for seen and unseen classes. 
 Red/Bold/Underlined text indicates best/2nd best/3rd best performance.
   }
        \end{threeparttable}

        \end{minipage}}
   \end{table*}

\begin{table*}[h!]
	%\small
    
      \resizebox{0.95\textwidth}{!}{\begin{minipage}{\textwidth} 
      \centering
    \begin{tabular}{|l|c|c|r|l|}

\hline
\textbf{KG} &   \textbf{N}  &  \textbf{E}   &  \textbf{RT} & \textbf{RC}\\ \hline\hline

% WN\_H2 & 70 / 54 / 49 &  321 / 223 / 105 &       5 &LX\\ \hline
% WN\_H3 & 429 / 311 / 295 &  873 / 680 / 655 &       5 &LX\\ \hline\hline
WN\_H2 & 70 / 54 / 49  / 79&  321 / 223 / 105 / 365 &       5 &LX\\ \hline
WN\_H3 & 429 / 311 / 295 / 465&  873 / 680 / 655 / 912 &       5 &LX\\ \hline\hline

CN\_H2 &   715 / 552 / 504 / 743 / & 2,132 / 1,981 / 1,864  / 2,342 &    13 &CS    \\ \hline
CN\_H3 & 2,139 / 1,872 / 1,788 /2,349 /  &  2,542 / 2,194 / 2,103 / 2,874 &       24 &CS    \\ \hline 
CN\_H2\_TH & 611 / 505 / 485  / 785 & 1,710 / 1,521 / 1,415 / 1,956 &       12 &CS  \\ \hline
CN\_H3\_TH & 12,733 / 9,839 / 9,212  / 13,045&  29,794 / 25,105 / 24,292 / 32,456 &       29 & CS \\ \hline \hline

CN+WN\_H2 & 667 / 581 / 506 / 845 & 1,906 / 1,682 / 1,602 / 2,136 &       13 &CS \\ \hline
CN+WN\_H2\_TH & 590 / 492 / 431 / 705 & 1,442 / 1,167 / 1,089 / 1,673&       12 &CS/LX    \\ \hline
CN+WN\_H3\_TH &  10,165 / 8,842 / 7,948 / 12,116 &  26,735 / 23,176 / 22,602 / 29,672  &       29 & CS/LX   \\ \hline 
  \end{tabular}

     % \end{minipage}}
  
   \caption{KGs Details. N: Number of Nodes. E: Number of Edges.  RT: Number of Different Relation Types between nodes. RC: Category of Relation Types. CS: Common-Sense. LX: Lexicographic. First/Second/Third/Fourth number in the N and E columns refers to the KG for OSDD/CGQA-States/MIT-States/ VAW dataset, respectively.}

    \label{tab:KGS}
     \end{minipage}}  
\end{table*}

\begin{table*}[h!] 
    \centering
     % \resizebox{0.90\textwidth}{!}{\begin{minipage}{\textwidth} 
\resizebox{\textwidth}{!}{     
    \begin{tabular}{|l|cccc|cccc|cccc|cccc|}

\hline
 \bf \multirow{2}{*}{Method}& \multicolumn{4}{|c|}{\textbf{OSDD}}  & \multicolumn{4}{|c|}{\textbf{CGQA-States}}  & \multicolumn{4}{|c|}{\textbf{MIT-States}}   & \multicolumn{4}{|c|}{\textbf{VAW}}  \\      \cline{2-17} 

\cline{2-13} 

& \textbf{S} & \textbf{Un}   & \textbf{HM} & \textbf{AUC} & \textbf{S} & \textbf{UN}   & \textbf{HM} & \textbf{AUC}
& \textbf{S} &\textbf{UN}  & \textbf{HM} & \textbf{AUC}
& \textbf{S} &\textbf{UN}  & \textbf{HM} & \textbf{AUC}
 \\ \hline \hline

 %\parbox[t]{2mm}{\multirow{12}{*}{\rotatebox[origin=c]{90}{GN}}}
 CN\_H3\_LSTM & 85.1 & 38.0 & 38.0 & 24.3 & 96.4 & 57.1 & 37.3 & 27.0 & 92.9 & 65.4 & 50.9 & 36.9 
&55.7 
&43.9 
&22.1 
&12.5 
 \\ 
CN\_H3\_GCN & 86.7 & 58.5 & \color{blue}\textbf{44.1} & \color{blue}\textbf{34.0}   & 95.7 & 62.5 & 40.0 & 28.7 & 88.1 & 66.7 & 47.1 & 32.2 

&70.3 
&49.5 
&30.2 
&20.8 

\\ 
CN\_H3\_R-GCN & 87.7 & 49.0 & 42.7 & 30.4 & 95.7 & 71.4 &\color{blue}\textbf{40.9}  & \color{blue} \textbf{34.0} & 78.6 & 73.4 &  47.4 & 32.9 
&79.5 
&57.5 
&38.9 
&28.8 
\\ 
 CN\_H3\_Tr-GCN & 87.4 & 42.2 & 40.2 & 27.7 & 93.6 & 56.3 & 39.2 & 28.8 & 88.1 & 67.0 &  \textbf{53.6} & \textbf{43.7} 
 &80.2 
&56.8 
&\color{blue} \textbf{40.7} 
& \color{blue} \textbf{29.9} 
 \\ 

\hline
WN\_H3\_LSTM & 86.0 & 60.0 & \color{blue} \textbf{43.3} & \color{blue} \textbf{33.9} & 96.4 & 13.4 & 16.6 & 8.7 & 90.5 & 24.4 & 24.2 & 13.2 
&37.4 &55.6 & 18.1 & 10.2 

\\ 
WN\_H3\_GCN & 86.8 & 39.5 & 36.7 & 21.2 & 86.4 & 49.0 & 34.2 & 24.1 & 88.1 & 54.8 & \color{blue} \textbf{50.1} & \color{blue} \textbf{37.9} 
&64.2 &38.3 & 24.4 &19.4

\\ 
WN\_H3\_R-GCN & 85.5 & 36.0 & 36.5 & 22.1 & 93.6 & 52.9 & \color{blue} \textbf{40.5} &  \color{blue}\textbf{28.9} & 78.6 & 47.4 & 42.9 & 21.4 

&69.7 &56.0 & 38.9 &28.8

\\ 
WN\_H3\_Tr-GCN & 89.2 & 48.4 & 36.6 & 23.9 & 86.4 & 56.6 & 37.6 & 26.6 & 88.1 & 44.2 & 37.3 & 25.9
& 65.0 & 54.5 & \color{blue} \textbf{31.8} & \color{blue} \textbf{21.3}
\\ 

\hline

CN\_H2\_TH\_LSTM & 86.5 & 50.0 & 43.0 & 28.8 & 97.1 & 71.7 & 38.8 & 31.9 & 78.6 & 60.3 & 47.8 & 26.0 

& 61.0 & 52.6 & 27.9 & 17.9 

\\ 
%CN+WN\_H2\_TH\_GCN
CN\_H2\_TH\_GCN & 84.6 & 52.8 & 43.7 & 30.7 & 95.7 & 67.5 & 40.5 & 32.0 & 85.7 & 73.1 & 46.6 & 29.4 

 & 74.3 & 48.3 & 36.4 & 27.4

\\ 
 %CN+WN\_H2\_TH\_R-GCN
CN\_H2\_TH\_R-GCN & 85.9 & 48.0 & 41.2 & 28.5 & 95.0 & 63.6 & 41.6 & 31.6 & 81.0 & 69.2 & 51.8 & 30.0

&82.4 
&57.6 
& 40.5
& 31.5 
\\

CN\_H2\_TH\_Tr-GCN & 85.7 & 63.7 & \textbf{45.6} & \textbf{34.5} & 97.1 & 70.0 &\textbf{43.5} & \textbf{35.6} & 85.7 & 70.2 & \color{blue} \textbf{51.6} & \color{blue} \textbf{40.5}   
% &82.4 &59.4 &38.0 & 32.6
&82.4&59.4 &\textbf{38.0}  & \textbf{32.6}  \\
\hline
\hline\hline

WN\_H2\_Tr-GCN & 87.9 & 23.0 & 28.6 & 13.0 & 92.9 & 53.8 & \color{blue}\textbf{38.2} & \color{blue}\textbf{28.1} & 83.3 & 45.8 & \color{blue} \textbf{39.7} & \color{blue}\textbf{27.3}  

& 69.7 & 45.8 & 30.5 & 18.3 \\
WN\_H3\_Tr-GCN & 89.2 & 48.4 & \color{blue}\textbf{36.6} & \color{blue}\textbf{23.9} & 86.4 & 56.6 & 37.6 & 26.6 & 88.1 & 44.2 & 37.3 & 25.9 
& 65.0 & 54.5 & \color{blue}  \textbf{31.8}
& \color{blue}  \textbf{21.3}

\\ \hline

CN\_H2\_Tr-GCN & 86.4 & 60.6 & \color{blue}\textbf{45.1} & \color{blue} \textbf{34.3} & 97.1 & 73.4 &  \textbf{46.3} & \textbf{39.5} & 88.1 & 69.6 & \textbf{56.2} & \textbf{43.5} 
&82.4 
&58.9 
&\color{blue}\textbf{37.3}
&\color{blue}\textbf{32.0}
\\ 
CN\_H3\_Tr-GCN & 87.4 & 42.2 & 40.2 & 27.7 & 93.6 & 56.3 & 39.2 & 28.8 & 88.1 & 67.0 & 53.6 & 43.7 
&81.1 
&48.3 
&36.9
&26.3
 
\\ \hline

CN\_H3\_UN\_Tr-GCN & 85.7 & 14.8 &\color{blue} \textbf{17.0} & \color{blue}\textbf{7.6} & 93.6 & 13.2 & \color{blue} \textbf{15.1} &\color{blue} \textbf{7.4} & 83.3 & 26.6 & \color{blue}\textbf{20.6} & \color{blue} \textbf{7.6} 
&83.1 
&10.2 
&\color{blue} \textbf{14.8} 
&\color{blue} \textbf{5.3}

\\  
RN\_Tr-GCN & 12.9 & 11.3 & 3.2 & 1.6 & 15.7 & 9.7 & 5.1 & 2.5 & 26.7 & 24.2 & 12.5 & 4.6   
& 12.0 & 9.8 & 3.0 & 1.3 \\
\hline
CN+WN\_H2\_Tr-GCN & 85.7 & 60.9 & 45.2 & 33.9 & 97.1 & 72.0 & \color{blue}\textbf{46.0} & \color{blue} \textbf{38.9} & 88.1 & 68.9 & 55.3 & 43.3 
&82.0 
&58.9 
&39.8 
&32.6

\\ 
CN+WN\_H2\_TH\_Tr-GCN  

& 87.7 & 69.9 & \bf  \textcolor{black}{48.6} & \bf \textcolor{black} {39.8} & 
  97.1 & 73.4 &  43.6 & {36.5} & 
  85.7 & 69.9 & \bf  \textcolor{blue}{51.1} & \bf  \textcolor{blue}{41.2} 
  &83.7
&58.6 
& \textbf{42.9} 
& \textbf{32.8}

  \\
\hline\hline
\hline
 WN\_H2\_Tr-GCN  & 87.9 & 23.0 & 28.6 & 13.0 & 92.9 & 53.8 & \color{blue}\textbf{38.2} & \color{blue} \textbf{28.1} & 83.3 & 45.8 & \color{blue} \textbf{39.7} & \color{blue}\textbf{27.3}  
& 69.7 & 45.8 & 30.5 & 18.3 \\
  WN\_H3\_Tr-GCN & 89.2 & 48.4 & \color{blue} \textbf{36.6} &\color{blue}  \textbf{23.9} & 86.4 & 56.6 & 37.6 & 26.6 & 88.1 & 44.2 & 37.3 & 25.9 
& 65.0 & 54.5 & \color{blue}  \textbf{31.8} 
& \color{blue}  \textbf{21.3}\\
  
  \hline
CN\_H2\_Tr-GCN & 86.4 & 60.6 & 45.1 & 34.3 & 97.1 & 73.4 &  \textbf{46.3} &  \textbf{39.5} & 88.1 & 69.6 & \textbf{56.2} & \textbf{43.5} 
&82.4 
&58.9 
&\color{blue}\textbf{37.3}
&\color{blue}\textbf{32.0}

\\ 
CN\_H3\_Tr-GCN & 87.4 & 42.2 & \color{blue} \textbf{40.2} &  \color{blue}\textbf{27.7} & 93.6 & 56.3 & 39.2 & 28.8 & 88.1 & 67.0 & 
53.6 & 43.7 
&80.2 
&56.8 
&40.7
&29.9

\\  \hline
CN+WN\_H2\_TH\_Tr-GCN &  87.7 & 69.9 & \bf  \textcolor{black}{48.6} & \bf \textcolor{black} {39.8} & 
  97.1 & 73.4 &  \textcolor{blue}{43.6} & \textcolor{blue}{36.5} & 
  85.7 & 69.9 &  {51.1} &  {41.2} 
  
  &83.7
&58.6 
&\color{black}\textbf{42.9 }
&\color{black}\textbf{32.8 } 
  \\

CN+WN\_H3\_TH\_Tr-GCN & 87.1 & 56.3 & 44.6 & 31.9 & 97.1 & 60.5 & 41.0 & 32.5 & 83.3 & 68.6 & \color{blue} \textbf{55.9} & \color{blue} \textbf{41.0}   
&80.6 
&59.2 
&38.8 
&30.6  \\
\hline
  \hline\hline

WN\_H3\_Tr-GCN & 87.3 & 46.4 &  \color{blue} \textbf{35.7} & 23.0 & 85.5 & 53.6 & 35.3 & 25.2 & 87.2 & 44.3 &  \color{blue} \textbf{37.4} & 25.7 
& 65.0 & 54.5 & 31.8 & 21.3\\ 
WN\_H3\_TH\_Tr-GCN & 89.2 & 48.4 & 36.6 &  \color{blue} \textbf{23.9} & 86.4 & 56.6 &  \color{blue} \textbf{37.6} &  \color{blue} \textbf{26.6} & 88.1 & 44.2 & 37.3 &  \color{blue} \textbf{25.9} 
& 68.1 & 56.0 &  \color{blue} \textbf{32.7} &  \color{blue} \textbf{23.4}\\   \hline
CN\_H2\_Tr-GCN & 86.4 & 60.6 & 45.1 & 34.3 & 97.1 & 73.4 & \textbf{46.3} & \textbf{39.5} & 88.1 & 69.6 & \textbf{56.2} & \textbf{43.5} 
&82.4 
&58.9 
&37.3
&32.0
\\   
CN\_H2\_TH\_Tr-GCN & 85.7 & 63.7 & \textbf{45.6} & \textbf{34.5} & 97.1 & 70.0 & 43.5 & 35.6 & 85.7 & 70.2 & 51.6 & 40.5 
&82.4&59.4 &\textbf{38.0}  & \textbf{32.6}  
 \\

\hline 

  \end{tabular}

}  

\caption{Ablation Study. 
1st  section of the table: comparison for the GNN architecture. 
2nd section: comparison for the KG source. 
3rd section: comparison for max number of hops. 
4th section: comparison for the node inclusion policy. 
Bold font indicates top performance across ablation category. Blue colour indicates top performance across ablation subcategory.
S: Best Accuracy on seen classes. UN: Best accuracy on unseen classes. HM: Best harmonic mean. AUC: Area under curve for the  pairs of accuracy for seen and unseen classes.  CN: ConceptNet-based model. WN: WordNet-based model. UN: Embeddings corresponding to concepts unrelated to the target classes. RN: Random embeddings. H2(3): Maximum number of hops equal to 2(3). TH: Thresholding policy for the nodes of the KG.}
     % \end{minipage}}
   \label{tab:ablation}
   
\end{table*}

\end{document}